\documentclass[pdflatex,sn-mathphys-num]{sn-jnl}

\usepackage{graphicx}%
\usepackage{multirow}%
\usepackage{amsmath,amssymb,amsfonts}%
\usepackage{amsthm}%
\usepackage{mathrsfs}%
\usepackage[title]{appendix}%
\usepackage{xcolor}%
\usepackage{textcomp}%
\usepackage{subcaption} %
\usepackage{manyfoot}%
\usepackage{booktabs}%
\usepackage{algorithm}%
\usepackage{algorithmicx}%
\usepackage{algpseudocode}%
\usepackage{listings}%
\usepackage{tcolorbox}%
\usepackage{fvextra}

\newcommand{\modcolor}{black}
\newcommand*{\modified}[1]{\textcolor{\modcolor}{#1}}
\newcommand{\modcolorsnd}{black}
\newcommand*{\modifiedsnd}[1]{\textcolor{\modcolorsnd}{#1}}

\theoremstyle{thmstyleone}%
\theoremstyle{thmstyletwo}%

\theoremstyle{thmstylethree}%

\begin{document}

\title{An Evolutionary Framework for Automatic Optimization Benchmark Generation via Large Language Models}


\author[1]{\fnm{Yuhiro} \sur{Ono}}\email{ono-yuhiro@ed.tmu.ac.jp}

\author*[2]{\fnm{Tomohiro} \sur{Harada}}\email{tharada@mail.saitama-u.ac.jp}

\author[1]{\fnm{Yukiya} \sur{Miura}}\email{miura@tmu.ac.jp}

\affil[1]{\orgdiv{Graduate School of Systems Design}, \orgname{Tokyo Metropolitan University}, \orgaddress{\street{6-6, Asahigaoka}, \city{Hino}, \postcode{1910065}, \state{Tokyo}, \country{Japan}}}

\affil*[2]{\orgdiv{Graduate School of Science and Engineering}, \orgname{Saitama University}, \orgaddress{\street{255 Shimo-okubo, Sakura-ku}, \city{Saitama}, \postcode{3388570}, \state{Saitama}, \country{Japan}}}


\abstract{Optimization benchmarks play a fundamental role in assessing algorithm performance; however, existing artificial benchmarks often fail to capture the diversity and irregularity of real-world problem structures, while benchmarks derived from real-world problems are costly and difficult to construct. To address these challenges, we propose an evolutionary automatic benchmark generation framework that leverages a large language model (LLM) as a generative operator, termed the LLM-driven evolutionary benchmark generator (LLM-EBG). In this framework, the LLM serves as an evolutionary operator that generates and evolves benchmark problems within a flexible, expressive representation space. As a case study, we generate unconstrained single-objective continuous minimization problems represented as mathematical expressions designed to induce significant performance differences between a genetic algorithm (GA) and differential evolution (DE). Experimental results show that LLM-EBG successfully produces benchmark problems in which the designated target algorithm consistently outperforms the comparative algorithm in more than 80\% of trials. Furthermore, exploratory landscape analysis reveals that benchmarks favoring GA are highly sensitive to variable scaling, demonstrating that the proposed framework can generate problems with distinct geometric characteristics that reflect the intrinsic search behaviors of different optimization algorithms.}

\keywords{Evolutionary computation, Large language models, Generative AI-driven optimization, Automatic benchmark generation, Benchmark evolution, Algorithm performance differentiation}



\maketitle

\section{Introduction}\label{sec:introduction}
Optimization benchmarks play a fundamental role in evaluating the efficiency, robustness, and solution quality of optimization algorithms~\cite{PIOTROWSKI2023101378,HELLWIG2019927}. Since algorithmic performance is inherently dependent on problem structure, as stated by the \modifiedsnd{No Free Lunch theorem}~\cite{David1997}, evaluations based on a limited or structurally biased set of benchmark problems may lead to misleading conclusions regarding algorithmic superiority. Meaningful assessment of optimization algorithms, therefore, requires benchmark suites that encompass diverse structural characteristics.

Most conventional optimization benchmarks are artificially constructed using predefined mathematical formulations, such as the COCO (COmparing Continuous Optimizers)~\cite{Hansen2021} and CEC~\cite{Ahari2022, Liang2019} suites for single-objective optimization, and the ZDT~\cite{Zitzler2000}, DTLZ~\cite{Deb2002}, WFG~\cite{Huband2006}, and MaF~\cite{Cheng2007} suites for multi-objective optimization. These benchmarks provide controlled problem characteristics such as multimodality, non-separability, and ill-conditioning, enabling systematic analysis of algorithmic behavior. However, their predefined and stylized constructions restrict the range of structural variations that can be represented. As a result, they may fail to capture heterogeneous and irregular structures commonly observed in real-world optimization problems~\cite{Tanabe2017, Zapotecas2019, Nan2024GECCO, Malan2025GECCO}, and performance observed on such benchmarks does not necessarily reflect algorithmic behavior under practical conditions.

Benchmarks derived from real-world optimization problems, such as RWCMOP~\cite{KUMAR2021100961}, Mazda CdMOBP~\cite{Kohira2018ProposalOB}, and RE problems~\cite{TANABE2020106078}, enable evaluation under realistic conditions. However, they are often computationally expensive, difficult to construct, and restricted from public release due to confidentiality and intellectual property constraints. These limitations hinder reproducibility and scalability of algorithmic evaluation.

These limitations motivate the need for automatic benchmark generation methods that can produce problem instances with desired structural characteristics while maintaining reproducibility and manageable evaluation cost. However, existing automatic benchmark generation approaches typically rely on predefined parametric generators~\cite{Lou2018}, evolutionary instance generators tailored to specific problem formulations~\cite{Lechien2023}, or tree-based symbolic representations such as genetic programming (GP)~\cite{He2024,Seiler2025RandOptGen}. 
\modified{These approaches share a common limitation: the generated benchmark problems are confined to the space defined by their predefined structural constraints. For example, parametric generators can only produce problems within the range of their tunable parameters, evolutionary instance generators are tailored to specific problem formulations and lack generalizability, and GP-based approaches require the designer to predefine the tree structure along with the terminal and non-terminal symbols, such as operators, constants, and variables, that constitute the search space. Therefore, these} representations require prior specification of benchmark structures and consequently constrain the diversity and expressiveness of the generated problem landscapes.

To address these limitations, our preliminary conference paper~\cite{Ono2025GECCO} proposed an evolutionary automatic benchmark generation framework that leverages a large language model (LLM) as a generative operator, termed the LLM-driven evolutionary benchmark generator (LLM-EBG).
While the initial study introduced the basic concept of LLM-driven benchmark evolution, the present work substantially extends this framework by systematically consolidating its algorithmic formulation and deepening the structural analysis of the generated benchmarks, thereby establishing a more principled and reliable benchmark generation methodology.
The proposed method constructs benchmark problems as text-based symbolic expressions, while evolutionary objectives guide the search toward benchmarks exhibiting desired structural characteristics.

As a case study, experiments are conducted to generate unconstrained single-objective five-dimensional benchmark problems that induce performance disparities between two evolutionary algorithms, a genetic algorithm (GA)~\cite{Holland1975} and differential evolution (DE)~\cite{Storn1997}. \modified{Specifically, we consider a scenario in algorithm development where a practitioner seeks to understand the structural characteristics of problems that reveal the strengths and weaknesses of algorithms, with the goal of facilitating algorithm improvement and clarifying their applicability. To this end}, benchmark instances are generated in which GA outperforms DE, and vice versa. \modifiedsnd{To validate the effectiveness of LLM-EBG, its benchmark generation performance is compared with three baselines: HFEBG~\cite{Lou2018}, a conventional tree-based GP generator, and a random search. The comparison with GP and random search, in particular, isolates the contribution of the LLM by replacing the LLM-based operators while keeping the same representation and evaluation, thereby clarifying whether the LLM exploits the search history rather than acting as a random generator.}

The remainder of this paper is organized as follows.
Section~\ref{sec:related} reviews LLMs, their integration with EAs, and existing automatic benchmark generation approaches.
Section~\ref{sec:LLM-EBG} introduces the proposed LLM-EBG and details its algorithmic formulation.
Section~\ref{sec:experiment} presents the experimental settings and evaluation protocol.
Section~\ref{sec:result} reports the benchmark generation results and analyzes the landscape characteristics of the generated problems.
Section~\ref{sec:analysis_behavior} investigates the evolutionary dynamics of LLM-EBG by analyzing operator usage patterns and lineage structures.
\modified{Section~\ref{sec:limitation} discusses the limitations of the proposed method, including computational cost and scalability to higher-dimensional problems.}
Finally, Section~\ref{sec:conclusion} concludes the paper and outlines directions for future research.

\section{Related Work}\label{sec:related}
This section reviews large language models, their integration with evolutionary algorithms, and existing automatic benchmark generation methods.

\subsection{Large Language Models}
LLMs~\cite{Radford2018} are natural language processing models trained through self-supervised learning on large-scale corpora. Recent advances, such as instruction fine-tuning~\cite{Wei2022} and reinforcement learning from human feedback (RLHF)~\cite{Christiano2017, Stiennon2020}, have significantly enhanced their ability to perform complex reasoning, program synthesis, and structured content generation. These developments have stimulated increasing interest in incorporating LLMs into evolutionary computation frameworks.

\subsection{LLM as Evolutionary Operators}\label{sec:LLMandEC}
A growing body of research integrates LLMs with EAs to replace manually designed genetic operators. Representative methods include Language Model Crossover (LMX)~\cite{Meyerson2024}, EvoLLM~\cite{Lange2024}, and LLM-driven EA (LMEA)~\cite{Liu2024}. These approaches exploit the in-context learning (ICL) capability of LLMs~\cite{Brown2020} to infer structural patterns from parent individuals and generate offspring without explicit operator design. For example, LMX predicts offspring sequences from parent solutions, EvoLLM proposes new candidates based on archived high-fitness individuals, and LMEA performs crossover and mutation through natural-language instructions in combinatorial optimization tasks. Such LLM-based operators enable flexible, domain-agnostic generation of high-quality solutions.

Although LLMs inherently rely on next-token prediction and may exhibit tendencies toward local pattern repetition, embedding them within evolutionary loops enables global exploration through iterative evaluation and feedback. This interaction allows LLM-guided search processes to progressively steer solution generation toward promising regions of the search space.

Beyond solution generation, recent studies have explored evolutionary refinement of algorithms and code. Methods such as FunSearch~\cite{Romera2024}, EUREKA~\cite{Yecheng2024}, and LLaMEA~\cite{Stein2025} employ LLMs to iteratively generate, evaluate, and improve algorithmic components. These frameworks perform population-based exploration, mitigate premature convergence to specific code patterns, and have demonstrated the discovery of novel algorithmic structures through evolutionary search.

\subsection{Automatic Benchmark Generators}\label{sec:generator}
Several evolutionary approaches have been proposed for automatically generating benchmark problems that emphasize performance disparities among optimization algorithms.

The hierarchical-fitness-based evolving benchmark generator (HFEBG)~\cite{Lou2018} employs a parameterized benchmark generator whose problem characteristics are controlled through tunable parameters, and uses an evolutionary algorithm to search for parameter configurations that favor a target algorithm.

Lechien et al.~\cite{Lechien2023} proposed an evolutionary generation of Hamiltonian cycle problem instances to induce differences in computational time between solvers. He and Aranha~\cite{He2024} introduced a GP–based framework that evolves benchmark functions to maximize the discrepancy between solution distributions of two algorithms using the Wasserstein distance as the evaluation criterion.

In contrast to these approaches, which rely on predefined structural templates or random symbolic compositions, the proposed LLM-EBG introduces LLMs as generative operators to evolve benchmark problems directly in a symbolic expression space under explicitly defined algorithm-discriminative objectives, \modified{thereby enabling the generation of problems within a more flexible representational space (detailed in Section~\ref{sec:algorithm}).}

\section{LLM-EBG: LLM-driven Evolutionary Optimization Benchmark Generator}\label{sec:LLM-EBG}

This study introduces the LLM-driven evolutionary benchmark generator (LLM-EBG), an automatic benchmark generation framework that integrates EAs with LLMs. LLM-EBG treats benchmark problems as individuals in an evolutionary process and employs an LLM as a generative operator to construct and evolve benchmark instances represented by symbolic expressions. This section first provides an overview of LLM-EBG, followed by a description of the LLM-based benchmark generation process, the evaluation method for generated benchmarks, and the overall algorithmic procedure of LLM-EBG.

\subsection{Overview}\label{sec:algorithm}
\begin{figure}[tb]
\centering
\includegraphics[width=\linewidth]{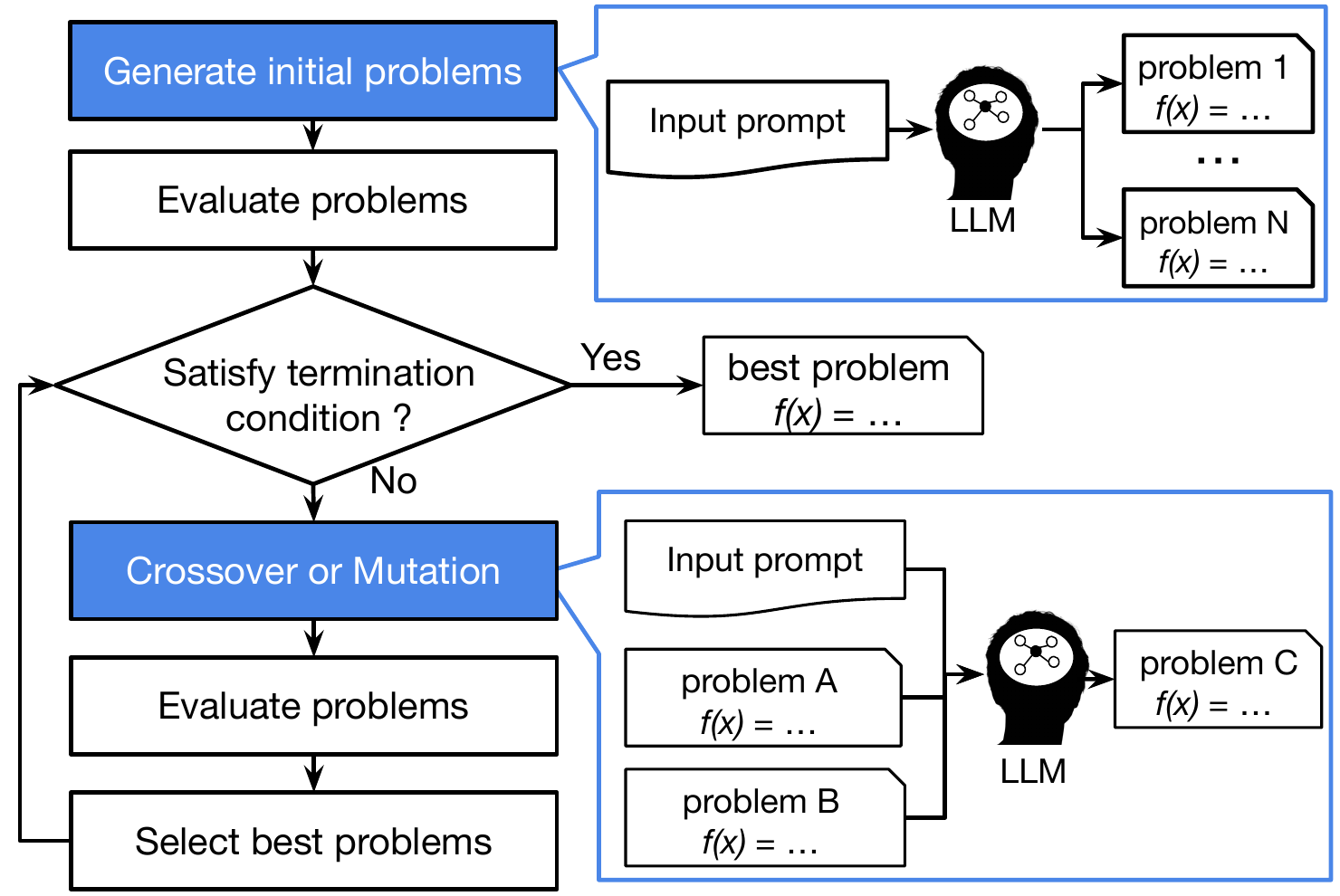}    
\caption{Workflow of LLM-EBG for benchmark generation}\label{fig:llm-ebg}
\end{figure}

Fig.~\ref{fig:llm-ebg} illustrates the workflow of LLM-EBG. The blue nodes indicate the steps in which LLM is invoked (detailed in Section~\ref{sec:operator}). In LLM-EBG, each benchmark problem is treated as an individual in an evolutionary population, and new benchmark instances are generated through LLM-based crossover and mutation operators. The generated benchmarks are evaluated according to predefined objective functions that quantify desired problem characteristics, such as algorithmic performance behavior and fitness landscape properties, and the resulting evaluation values are used to guide population updates and selection. Although various evaluation objectives can be defined within this framework, this study focuses on generating benchmarks that emphasize the performance disparities between two distinct optimization algorithms (detailed in Section~\ref{sec:evaluator}).

\modified{There are three key advantages of employing LLMs as generative operators in LLM-EBG. First, by leveraging few-shot prompting, LLMs can generate new symbolic expressions that are structurally similar to given examples, naturally realizing \modifiedsnd{crossover-like and mutation-like operations} in the symbolic expression space without requiring explicit operator design.
\modifiedsnd{Second, unlike existing approaches which are confined to predefined structural constraints---such as tunable parameter ranges, specific problem formulations, or tree-based representations with fixed terminal and non-terminal symbols---LLMs are not inherently bound to a specific representation and can, in principle, handle various forms of representations, such as mathematical expressions and program code. It should be noted that the current implementation of LLM-EBG also imposes a representational constraint, as benchmark problems are restricted to a single mathematical expression. Nevertheless, since this restriction stems from the experimental design of this study rather than from the framework itself, LLM-EBG can be positioned as a first step toward a more general framework capable of generating benchmarks in broader representational spaces.} Third, LLMs have acquired extensive mathematical knowledge through pre-training, enabling them to naturally generate expressions that avoid computationally infeasible regions, such as division by zero and complex numbers, without requiring explicit constraints.}

\subsection{Genetic Operators Using LLM}\label{sec:operator}
LLM-EBG employs LLM as a generative operator to construct and modify benchmark problems during the evolutionary process. In this framework, benchmark problems are represented as symbolic expressions, and the LLM is invoked to realize initialization, crossover, and mutation operators.

The output of the LLM is controlled through prompt-based instructions. Among representative prompt engineering strategies, zero-shot prompting and few-shot prompting~\cite{Brown2020} are commonly used. Zero-shot prompting generates outputs solely from a given query, whereas few-shot prompting conditions the model on a small number of examples provided within the prompt. In LLM-EBG, few-shot prompting is adopted to standardize the output format of generated benchmarks and to induce structural transformations required for genetic operations.

\begin{figure}[tb]
\begin{tcolorbox}
\begin{Verbatim}[breaklines=true, breakanywhere=true, fontsize=\footnotesize, breaksymbolleft=, breaksymbolright=]
You are an expert in generating optimization benchmark problems.
Create a new {D}-dimensional problem that {A1} outperforms {A2}.

{examples}

### Instructions ###
1. Generate one problem function `f(x)` in {D} dimensions.
2. Use only the following operators:[+,-,*,/,**,sqrt,sin,sinh,abs].
3. Write in a single line of Python code, starting with `Problem: f(x) = '.
4. Output only the required Python code line. Do not provide any explanation, preamble, or concluding remarks.

Problem: f(x) =
\end{Verbatim}
\end{tcolorbox}
\caption{The input template in the proposed method}
\label{fig::promptFormat}
\end{figure}

Fig.~\ref{fig::promptFormat} shows the prompt template used in LLM-EBG. The framework considers unconstrained single-objective continuous minimization problems represented by mathematical expressions. In the figure, the term ``\texttt{D}'' denotes the number of design variables. The terms ``\texttt{A1}'' and ``\texttt{A2}'' denote the optimization algorithms used for benchmark evaluation, where the generated benchmarks are required to induce a performance advantage of the algorithm A1 over the algorithm A2. The term ``\texttt{examples}'' refers to benchmark instances in the current population that are provided as in-context examples. The prompt specifies the target performance relation between algorithms A1 and A2 and constrains the function representation and output format.

The following subsections describe the realization of initialization, crossover, and mutation operators in LLM-EBG.

\subsubsection{Initialization}
\label{sec:init}
In the initialization phase, the LLM is used to generate an initial population of $N$ benchmark problems. The first individual is initialized using a predefined seed function \begin{quote}
    \modified{\texttt{x[0]+x[1]**2+x[2]**3$\cdots$+x[D]**(D-1)}}
\end{quote} 
and the subsequent $2$nd to $N$-th individuals are generated by conditioning the LLM on previously generated benchmarks as in-context examples. This procedure induces structural similarity among individuals while introducing new functional components.

For example, in Fig.~\ref{fig::promptInitialization}, three example benchmarks sharing the common subexpression \begin{quote}
    \modified{\texttt{x[0]**2 + sin(x[1])**2 + abs(x[2]*x[3]) + sqrt(abs(x[4])) + x[0]}} 
\end{quote}
are provided to the LLM, and a new benchmark that retains this subexpression while incorporating an additional term is generated.
\begin{figure}[tb]
\begin{tcolorbox}[title=Input]
\begin{Verbatim}[breaklines=true, breakanywhere=true, fontsize=\footnotesize, breaksymbolleft=, breaksymbolright=]
You are an expert in generating optimization benchmark problems.
Create a new 5-dimensional problem that GA outperforms DE.

Example 1:
f(x) = x[0]**2 + sin(x[1])**2 + abs(x[2]*x[3]) + sqrt(abs(x[4])) + x[0]*x[1]*x[2]
Example 2:
f(x) = x[0]**2 + sin(x[1])**2 + abs(x[2]*x[3]) + sqrt(abs(x[4])) + x[0]*sin(x[1])*abs(x[2])
Example 3:
f(x) = x[0]**2 + sin(x[1])**2 + abs(x[2]*x[3]) + sqrt(abs(x[4])) + x[0]*x[1]*sin(x[2])

### Instructions ###
1. Generate one problem function `f(x)` in 5 dimensions.
2. Use only the following operators:[+,-,*,/,**,sqrt,sin,sinh,abs].
3. Write in a single line of Python code, starting with `Problem: f(x) = '.
4. Output only the required Python code line. Do not provide any explanation, preamble, or concluding remarks.

Problem: f(x) =
\end{Verbatim}
\end{tcolorbox}
\begin{tcolorbox}[title=Output]
\begin{Verbatim}[breaklines=true, breakanywhere=true, fontsize=\footnotesize, breaksymbolleft=, breaksymbolright=]
x[0]**2 + sin(x[1])**2 + abs(x[2]*x[3]) + sqrt(abs(x[4])) + x[0]*sin(x[1])*x[2]*x[3]
\end{Verbatim}
\end{tcolorbox}
\caption{The input/output example for the initial population generation ($n=4$)}
\label{fig::promptInitialization}
\end{figure}

\subsubsection{Crossover}
\label{sec:crossover}
For crossover, two benchmark problems are randomly selected from the current population and provided to the LLM as in-context examples. 
\modifiedsnd{The two parents are provided merely as in-context examples, and whether and to what extent their components are reused is left to the LLM. Nevertheless, owing to the tendency of few-shot prompting to produce outputs that follow the patterns of the provided examples, the generated benchmark typically inherits structural components of the parents. In this sense, providing two parents as in-context examples is expected to realize a crossover-like operation.}

\begin{figure}[tb]
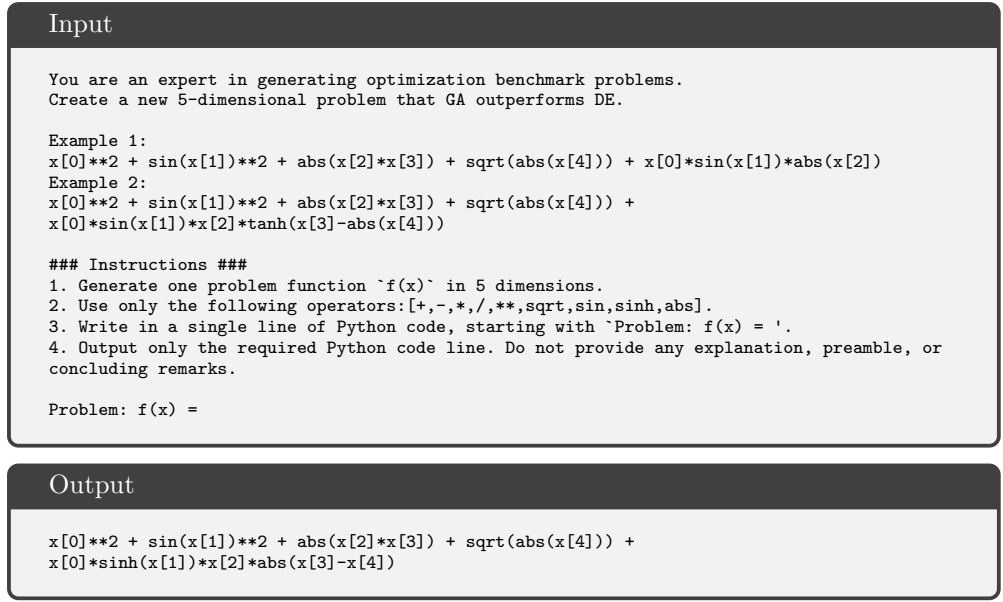

\begin{tcolorbox}[title=Input]
\begin{Verbatim}[breaklines=true, breakanywhere=true, fontsize=\footnotesize, breaksymbolleft=, breaksymbolright=]
You are an expert in generating optimization benchmark problems.
Create a new 5-dimensional problem that GA outperforms DE.

Example 1:
x[0]**2 + sin(x[1])**2 + abs(x[2]*x[3]) + sqrt(abs(x[4])) + x[0]*sin(x[1])*abs(x[2])
Example 2:
x[0]**2 + sin(x[1])**2 + abs(x[2]*x[3]) + sqrt(abs(x[4])) + x[0]*sin(x[1])*x[2]*tanh(x[3]-abs(x[4]))

### Instructions ###
1. Generate one problem function `f(x)` in 5 dimensions.
2. Use only the following operators:[+,-,*,/,**,sqrt,sin,sinh,abs].
3. Write in a single line of Python code, starting with `Problem: f(x) = '.
4. Output only the required Python code line. Do not provide any explanation, preamble, or concluding remarks.

Problem: f(x) =
\end{Verbatim}
\end{tcolorbox}

\begin{tcolorbox}[title=Output]
\begin{Verbatim}[breaklines=true, breakanywhere=true, fontsize=\footnotesize, breaksymbolleft=, breaksymbolright=]
x[0]**2 + sin(x[1])**2 + abs(x[2]*x[3]) + sqrt(abs(x[4])) + x[0]*sinh(x[1])*x[2]*abs(x[3]-x[4])
\end{Verbatim}
\end{tcolorbox}

\caption{The input/output example for crossover}
\label{fig::promptCrossover}
\end{figure}

For example, in Fig.~\ref{fig::promptCrossover}, two parent benchmarks are provided to the LLM, and a new benchmark that partially contains components of both parents is generated.

\subsubsection{Mutation}
\label{sec:mutation}
For mutation, a single benchmark is randomly selected from the current population and provided to the LLM as an in-context example. 
\modifiedsnd{As with crossover, the single parent is provided merely as an in-context example, and whether and how it is modified is left to the LLM. Nevertheless, as the output tends to follow the pattern of the provided example, the generated benchmark typically corresponds to a modified version of the parent. In this sense, providing a single parent as an in-context example is expected to realize a mutation-like operation. The distinction between the two operators therefore lies solely in the number of in-context examples provided: two for crossover and one for mutation.}

\begin{figure}[tb]
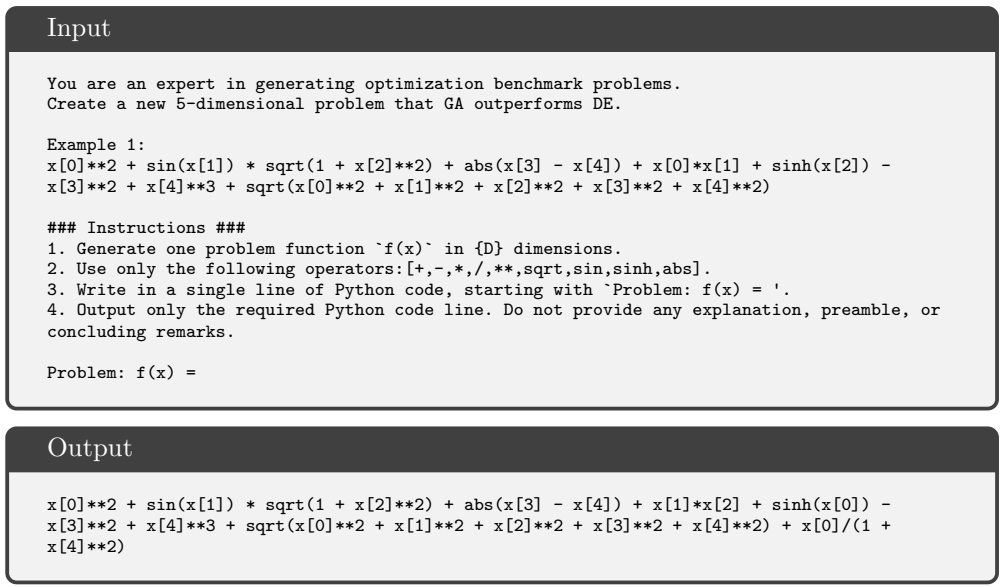

\begin{tcolorbox}[title=Input]
\begin{Verbatim}[breaklines=true, breakanywhere=true, fontsize=\footnotesize, breaksymbolleft=, breaksymbolright=]
You are an expert in generating optimization benchmark problems.
Create a new 5-dimensional problem that GA outperforms DE.

Example 1:
x[0]**2 + sin(x[1]) * sqrt(1 + x[2]**2) + abs(x[3] - x[4]) + x[0]*x[1] + sinh(x[2]) - x[3]**2 + x[4]**3 + sqrt(x[0]**2 + x[1]**2 + x[2]**2 + x[3]**2 + x[4]**2)

### Instructions ###
1. Generate one problem function `f(x)` in {D} dimensions.
2. Use only the following operators:[+,-,*,/,**,sqrt,sin,sinh,abs].
3. Write in a single line of Python code, starting with `Problem: f(x) = '.
4. Output only the required Python code line. Do not provide any explanation, preamble, or concluding remarks.

Problem: f(x) =
\end{Verbatim}
\end{tcolorbox}
\begin{tcolorbox}[title=Output]
\begin{Verbatim}[breaklines=true, breakanywhere=false, fontsize=\footnotesize, breaksymbolleft=, breaksymbolright=]
x[0]**2 + sin(x[1]) * sqrt(1 + x[2]**2) + abs(x[3] - x[4]) + x[1]*x[2] + sinh(x[0]) - x[3]**2 + x[4]**3 + sqrt(x[0]**2 + x[1]**2 + x[2]**2 + x[3]**2 + x[4]**2) + x[0]/(1 + x[4]**2)
\end{Verbatim}
\end{tcolorbox}
\caption{The input/output example for mutation}
\label{fig::promptMutation}
\end{figure}
For example, in Fig.~\ref{fig::promptMutation}, the benchmark
\begin{quote}
\modified{\texttt{f(x) = x[0]**2 + sin(x[1]) * sqrt(1 + x[2]**2) + abs(x[3] - x[4]) + x[0]*x[1] + sinh(x[2]) - x[3]**2 + x[4]**3 + sqrt(x[0]**2 + x[1]**2 + x[2]**2 + x[3]**2 + x[4]**2)}}
\end{quote}
is provided to the LLM, and a new benchmark by modifying the variables and introducing an additional term,
``\texttt{x[0]/(1 + x[4]**2)}'',
is generated.

\subsection{Evaluation of Generated Benchmarks}\label{sec:evaluator}
In LLM-EBG, the definition of desirable benchmark problems is specified through the evaluation function used in the evolutionary process. By changing the evaluation criteria, LLM-EBG can generate benchmark problems that emphasize different algorithmic properties. \modified{In this study, we consider a scenario in algorithm development where a practitioner seeks to identify the structural characteristics of problems that lead to performance differences among algorithms. In such a scenario, analyzing the generated benchmarks can provide insights into the strengths and weaknesses of each algorithm, thereby facilitating algorithm improvement and clarifying their applicability to specific problem classes. To this end, we design an evaluation function that generates benchmark problems in which the performance difference between a target algorithm (A1) and a comparative algorithm (A2) is pronounced.}

Specifically, benchmark problems are evaluated by the relative performance of the two algorithms over $T$ repeated trials, with smaller fitness values indicating stronger performance of A1 relative to A2. \modified{
Since the scale of the objective values of the generated benchmark problems is unspecified, a rank-based metric inspired by the Wilcoxon rank-sum test is employed for evaluation, as it enables performance comparison without assuming a specific distribution of objective values.} The evaluation function is defined as follows:
\color{\modcolor}
\begin{equation}
    fitness=\dfrac{\sum^{T}_{i=1} \text{rank}(q_{A1}^{i})}{\sum^{2T}_{n=1}n}+\alpha\times\max(0,-\min_i q^i_{A1}),
    \label{eq:evalPRC}
\end{equation}
\color{black}
where $T$ denotes the number of trials for each optimization algorithm, $q_{A1}^i$ represents the best value found by algorithm A1 in the $i$-th trial, and $\text{rank}(q_{A1}^i)$ is the rank of $q_{A1}^i$ when all trials of both A1 and A2 are pooled and ranked together. 
\modified{$\min_i q^i_{A1}$ is the minimum value obtained by the target algorithm across all $T$ trials, and the second term, $\alpha \times \max(0,-\min_i q^i_{A1})$, is a penalty term introduced to prevent the optimal value of the benchmark from diverging toward negative infinity.} 
The parameter $\alpha$ is a hyperparameter that adjusts the weight of this penalty. This term favors benchmark problems whose optimal solutions remain close to zero on the positive side. 
\modified{
The first term of Eq.~\eqref{eq:evalPRC} represents the proportion of the total rank sum accounted for by A1, where the denominator $\sum_{n=1}^{2T} n$ corresponds to the total rank sum across all $2T$ trials of both algorithms. When A1 consistently outperforms A2 across all $T$ trials, A1 occupies the lower $T$ ranks, and the first term attains its theoretical minimum, which serves as a clear and interpretable threshold for evaluating benchmark generation performance.
}

\modified{
Note that if at least one trial of either A1 or A2 produces invalid results (i.e., infinite or complex values), the benchmark is assigned a penalty fitness value of a significantly large positive value.}

\subsection{Overall Algorithm of LLM-EBG}
\begin{algorithm}[tb]
\caption{Pseudocode of LLM-EBG}
\label{alg::LLM-EBG}
\begin{algorithmic}[1]
\Require $N$: population size, $g_{\max}$: maximum number of generations, $p_c$: crossover rate
\Ensure Best benchmark in $P_{g_{\max}}$
\State Generate the initial population $P_{0}$ consisting of $N$ benchmarks
\State Evaluate each benchmark in $P_{0}$
\For{$g = 1$ to $g_{\max}-1$}
    \State $Q_g \gets \emptyset$
    \While{$|Q_g| < N$}
        \If{$\text{rand}(0,1) < p_c$}
            \State Randomly select two parents $p_1$ and $p_2$ from $P_{g-1}$
            \State Generate offspring $c$ by applying crossover to $p_1$ and $p_2$
        \Else
            \State Randomly select a parent $p$ from $P_{g-1}$
            \State Generate offspring $c$ by applying mutation to $p$
        \EndIf
        \If{$c$ is valid}
            \State $Q_g \gets Q_g \cup \{c\}$
        \EndIf
    \EndWhile
    \State Evaluate each benchmark in $Q_g$
    \State Select the top $N$ benchmarks as the next population $P_g$ from $P_{g-1} \cup Q_g$
\EndFor
\State \textbf{return} Best benchmark in $P_{g_{\max}}$
\end{algorithmic}
\end{algorithm}
Algorithm~\ref{alg::LLM-EBG} presents the overall procedure of LLM-EBG. In Algorithm~\ref{alg::LLM-EBG}, $N$ denotes the population size, $g_{\max}$ denotes the maximum number of generations, and $p_c$ denotes the crossover rate. LLM-EBG first generates an initial population of $N$ benchmark problems using the LLM (line~1), as described in Section~\ref{sec:init}, and evaluates them using the fitness function defined in Section~\ref{sec:evaluator} (line~2). The evolutionary process is then iterated for $g_{\max}-1$ generations (lines~3--19).

In each generation, offspring benchmarks are generated through either crossover (Section~\ref{sec:crossover}) or mutation (Section~\ref{sec:mutation}) using the LLM (lines~6--12). For crossover, two parent benchmarks are randomly selected from the current population and combined by the LLM to generate an offspring. For mutation, a single parent is randomly selected and modified by the LLM to generate an offspring. The crossover rate $p_c$ controls which operator is applied: crossover is applied with probability $p_c$, whereas mutation is applied with probability $1-p_c$. The next population is formed by selecting the top $N$ benchmark problems from the union of the current and offspring populations based on their fitness values (line 18).

As a pre-validation step (Section~\ref{sec:evaluator}), after each benchmark is generated (lines~1 and 13), $1000$ randomly sampled design variables are evaluated to verify that the benchmark produces valid real-valued outputs. If invalid values are observed, the benchmark is discarded and regenerated, and the corresponding parent individuals are reselected.

\section{Experimental Setting}\label{sec:experiment}
This section describes the experimental scenarios, parameter settings, and evaluation protocol used to assess the proposed LLM-EBG framework.

\subsection{Experimental Scenarios}
To investigate the effectiveness of LLM-EBG, we conduct experiments to generate benchmark problems that induce performance differences between a genetic algorithm (GA)~\cite{Holland1975} and differential evolution (DE)~\cite{Storn1997}. The target benchmarks are unconstrained single-objective continuous minimization problems defined over the search space $[-1,1]^5$.

Two scenarios are considered:
\begin{description}
\item[\textbf{GA-preferred search}:] GA is treated as the target algorithm (A1) and DE as the comparative algorithm (A2).
\item[\textbf{DE-preferred search}:] DE is treated as the target algorithm (A1) and GA as the comparative algorithm (A2).
\end{description}
Ten independent trials are performed for each scenario.

\subsection{Parameter Settings}
\label{sec:params}
\begin{table}[tb]
\caption{Parameter settings of LLM-EBG, GA, and DE}
\label{tab:exp_param}
\centering
\begin{tabular}{llr}
\toprule
Category & Parameter & Value \\
\midrule
\multirow{5}{*}{LLM-EBG} & Population size $N$ & 10 \\
 & Max generations $g_{\max}$ & 20 \\
 & Crossover rate $p_c$ & 0.5 \\
 & Penalty weight $\alpha$ & 10 \\
 & LLM model & Llama 3.3--70B \\
\midrule
\multirow{6}{*}{GA} & Population size & 50 \\
 & Max generations & 1000 \\
 & SBX crossover rate & 0.8 \\
 & PM mutation rate & 0.1 \\
 & Distribution index (SBX, PM) & 20 \\
 & Survival strategy & $(\mu+\lambda)$ \\
\midrule
\multirow{5}{*}{DE} & Population size & 50 \\
 & Max generations & 1000 \\
 & Strategy & DE/1/rand/bin \\
 & Scaling factor $F$ & 1.0 \\
 & Crossover rate $CR$ & 0.8 \\
\bottomrule
\end{tabular}
\end{table}

The parameter settings for LLM-EBG and the evaluation algorithms are summarized in Table~\ref{tab:exp_param}. 

For LLM-EBG, the population size is set to $N=10$, the maximum number of generations to $g_{\max}=20$, and the crossover rate to $p_c=0.5$. The penalty weight $\alpha$ in Eq.~\eqref{eq:evalPRC} is set to $10$. Llama~3.3--70B~\cite{llama3.3modelcard} is used as the LLM for benchmark generation.

To evaluate the generated benchmarks, both GA and DE use a population size of $50$ and are executed for $1000$ generations. GA employs tournament selection, simulated binary crossover (SBX)~\cite{Deb1995}, polynomial mutation (PM)~\cite{Deb1996}, and a $(\mu+\lambda)$ survival strategy. The crossover rate is set to 0.8, the mutation rate to 0.1, and the distribution indices for both SBX and PM to $20$. DE employs the DE/1/rand/bin strategy with a scaling factor of $F=1.0$ and a crossover rate of $CR=0.8$.

\modified{The population size and maximum number of generations of LLM-EBG are set to smaller values than those typically used in conventional evolutionary algorithms. This is primarily due to the substantial computational cost of LLM-based benchmark generation, as each query requires significant time to generate a benchmark expression. The cost of evaluating each generated benchmark by running GA and DE for $T = 20$ trials each further increases the overall execution time. }

\subsection{Evaluation Protocol}
Each generated benchmark problem is evaluated using the fitness function defined in Eq.~\eqref{eq:evalPRC}. For each benchmark, both GA and DE are independently executed for $T=20$ trials, and the best objective value obtained in each trial is recorded. The fitness value is computed from the pooled ranking of the best-found solutions.

Under these settings, the minimum value of the fitness function in Eq.~\eqref{eq:evalPRC} is attained when the target algorithm outperforms the comparative algorithm in all trials, yielding $\sum_{n=1}^{20}n / \sum_{n=1}^{40}n \approx 0.256$.

\modifiedsnd{\subsection{Baseline Comparison}
To evaluate the effectiveness of LLM-EBG against baseline methods, we conduct benchmark generation using three baselines: HFEBG~\cite{Lou2018}, a tree-based GP generator, and a random search. HFEBG is selected because it is designed to generate benchmark problems that induce performance differences among optimization algorithms, which is consistent with the objective of LLM-EBG. The GP and random search baselines are introduced to isolate the contribution of the LLM: they replace the LLM-based operators while keeping the same symbolic-expression representation, evaluation function, and number of benchmark evaluations, thereby clarifying whether the LLM exploits the search history rather than acting as a random generator. For all baselines, 10 independent trials are conducted for each scenario, and the population size and the maximum number of generations are set to $N=10$ and $g_{max}=20$, respectively, which are identical to those used in LLM-EBG. Since these values are matched to LLM-EBG, all methods are compared under an identical number of benchmark evaluations ($N \times g_{max}$). We note that this evaluation budget is small relative to the settings typically adopted for GP; this point is discussed in relation to the results in Section~\ref{sec:llm-ebg-performance}.
}

\modifiedsnd{\subsubsection{HFEBG}
HFEBG searches the parameter space of the max-set of Gaussians (MSG) generator to obtain benchmark problems that favor a target algorithm. For HFEBG, the number of local optima is set to $n_0=30$. Note that the fitness values for HFEBG ignore negative penalties, representing the proportion of the total rank sum attributed to the target algorithm, as HFEBG does not have a mechanism to account for cases where the optimal value is negative. Since each benchmark generated by HFEBG is represented as the optimized parameters of the fixed Gaussian-mixture generator rather than as a symbolic expression, HFEBG benchmarks cannot be exemplified as compact formulas in the same way as those generated by LLM-EBG.}

\modifiedsnd{\subsubsection{GP}
The GP baseline replaces the LLM-based initialization, crossover, and mutation with a conventional tree-based GP implemented using DEAP~\cite{DEAP_JMLR2012}. To ensure a fair comparison, the primitive set is restricted to the same operators as those allowed in the LLM prompt, i.e., $\{+,-,\times,/,**,\sqrt{|\cdot| },\sin,\sinh,|\cdot|\}$, together with variables $x_0,\dots,x_{D-1}$ and integer constants. The initial population is generated by the ramped half-and-half method, crossover is performed by subtree crossover, and mutation is performed by uniform subtree mutation. As in LLM-EBG, parents are selected by binary tournament based on fitness, an offspring is accepted only if it yields real-valued objective values over the feasible region, and an elitist $(\mu+\lambda)$ selection is applied. Each generated tree is converted to a symbolic expression and evaluated by the same evaluation function (Eq.~\eqref{eq:evalPRC}) as LLM-EBG.}

\modifiedsnd{\subsubsection{Random Search}
The random search baseline removes the evolutionary mechanism entirely. It repeats, $N\times g_{max}$ times, the same procedure used to generate the first individual of the LLM-EBG initial population, in which the LLM is prompted with only a fixed seed function as an in-context example. Each candidate is thus generated independently of all previously generated or evaluated candidates, and no selection, crossover, or mutation is applied. The evaluation function and the total number of evaluations are identical to those of LLM-EBG.}

\section{Result}\label{sec:result}
This section reports the experimental results and analyzes the generated benchmark problems. The analysis focuses on the search behavior of the optimization algorithms and the structural characteristics of the generated benchmarks.

\subsection{Benchmark Generation Performance of LLM-EBG}
\label{sec:llm-ebg-performance}
\begin{table}[tb]
    \centering
    \caption{Best benchmarks obtained in each trial for GA-preferred and DE-preferred searches}
    \label{tb:resAllSearch}
        \begin{tabular}{cccccc}
        \toprule
        Scenario&Trial&$fitness$&Best GA value&Best DE value& Gen. to min.\footnotemark[1]\\
        \midrule
        \multirow{10}{*}{GA-preferred}&1&$\mathbf{0.256}$&$7.00\times10^{-4}$&$0.134$&$1$ \\
        &2&$0.259$&$-3.38\times10^{4}$&$0.177$&-($19$) \\
        &3&$\mathbf{0.256}$&$0.864$&$1.08$&$8$ \\
        &4&$\mathbf{0.256}$&$0.270$&$0.270$&$7$ \\
        &5&$\mathbf{0.256}$&$0.278$&$0.544$&$8$($2$) \\
        &6&$\mathbf{0.256}$&$3.21\times10^{-2}$&$0.677$&$12$ \\
        &7&$\mathbf{0.256}$&$2.65\times10^{-7}$&$1.81\times10^{-2}$&$3$ \\
        &8&$0.288$&$6.96\times10^{-4}$&$3.75\times10^{-21}$&-($2$)\\
        &9&$\mathbf{0.256}$&$1.27\times10^{-2}$&$0.518$&$6$ \\
        &10&$\mathbf{0.256}$&$3.20\times10^{-2}$&$0.469$&$13$ \\
        \midrule
        \multirow{10}{*}{DE-preferred}&1&$\mathbf{0.256}$&$2.43\times10^{-3}$&$6.32\times10^{-14}$&$8$ \\
        &2&$\mathbf{0.256}$&$0.222$&$0.176$&$11$($0$) \\
        &3&$\mathbf{0.256}$&$0.971$&$0.873$&$8$($1$) \\
        &4&$\mathbf{0.256}$&$0.161$&$0.161$&$8$($1$) \\
        &5&$\mathbf{0.256}$&$7.52\times10^{-4}$&$1.77\times10^{-23}$&$3$($0$) \\
        &6&$7.64$&$-0.736$&$-0.736$&-($0$) \\
        &7&$\mathbf{0.256}$&$6.24\times10^{-4}$&$0.00$&$0$ \\
        &8&$\mathbf{0.256}$&$0.919$&$0.913$&$4$ \\
        &9&$\mathbf{0.256}$&$8.22\times10^{-4}$&$2.52\times10^{-23}$&$6$ \\
        &10&$\mathbf{0.256}$&$0.148$&$0.148$&$8$($1$)\\
        \bottomrule
        \end{tabular}
\footnotetext[1]{Generation at which the theoretical minimum fitness value was achieved. The values in parentheses indicate the first generation in which the target algorithm outperformed the comparative algorithm in all trials, while the optimal value was still negative.}
\end{table}

Table~\ref{tb:resAllSearch} summarizes the benchmark problems generated in both the GA-preferred and DE-preferred searches. The table reports the best fitness values of the generated benchmarks, the best objective values obtained by each optimization algorithm, and the generation in which the fitness value first reached the theoretical minimum ($0.256$). \modified{It is noted that the values here are rounded to three significant figures.} The values in parentheses indicate the first generation in which the target algorithm outperformed the comparative algorithm across most trials while the optimal value of the benchmark was still negative. Fitness values that attain the theoretical minimum are highlighted in bold.

As shown in Table~\ref{tb:resAllSearch}, the fitness values attained the theoretical minimum in 8 out of 10 runs for the GA-preferred search and in 9 out of 10 runs for the DE-preferred search. These results demonstrate that LLM-EBG successfully generated benchmark problems that admit positive optimal solutions and on which the designated target algorithm consistently outperformed the comparative algorithm across all trials.

In all runs of both scenarios, LLM-EBG generated benchmark problems during the search process for which the target algorithm outperformed the comparative algorithm across all trials, even though their fitness values did not reach the theoretical minimum. These benchmark states, whose corresponding generations are indicated in parentheses in Table~\ref{tb:resAllSearch}, correspond to local optima under the evaluation function in Eq.~\eqref{eq:evalPRC}. This is because the optimal values of these benchmarks are negative, and their fitness values are therefore degraded by the penalty term in the evaluation function.

In several runs, LLM-EBG successfully escaped from these local optima and further generated benchmark problems whose optimal values became positive, thereby attaining the global optimum of the evaluation function. In contrast, in the second trial of the GA-preferred search, the search remained at a local optimum corresponding to a benchmark with a negative optimal value.

In addition, the DE-preferred search tended to identify generations exhibiting target dominance earlier than the GA-preferred search. On average, the GA-preferred search required 7.25 generations, whereas the DE-preferred search required 6.67 generations. This tendency is consistent with the generally superior performance of DE over GA in unconstrained single-objective continuous optimization problems.

\begin{table}[tb]
\color{\modcolorsnd}
\caption{\modifiedsnd{Comparison of LLM-EBG with the baseline methods}}
\label{tb:comparison}
\centering
\begin{tabular}{llcccc}
\toprule
Scenario & Metric & LLM-EBG & HFEBG & GP & Random \\
\midrule
\multirow{3}{*}{GA-preferred}
  & Success rate   & 8/10 & 4/10   & 0/10   & 3/10 \\
  & Median fitness & 0.256 & 0.261 & 0.488  & 0.276 \\
  & $p$-value      & ---   & 0.0482 & $<$0.01 & 0.0142 \\
  \midrule
\multirow{3}{*}{DE-preferred}
  & Success rate   & 9/10  & 0/10   & 10/10 & 10/10 \\
  & Median fitness & 0.256 & 0.354  & 0.256  & 0.256 \\
  & $p$-value      & ---   & $<$0.01 & 0.647  & 0.647 \\
\bottomrule
\end{tabular}
\color{black}
\end{table}
\color{\modcolorsnd}
Table~\ref{tb:comparison} summarizes a comparison of the benchmark generation performance between LLM-EBG and the three baseline methods. The detailed per-trial results for the baselines are presented in Appendix~\ref{sec:append}. 
 
In the GA-preferred scenario, LLM-EBG clearly outperforms all three baselines in terms of success rate and median fitness. A one-sided Wilcoxon rank-sum test confirms that LLM-EBG achieves significantly lower fitness values than every baseline ($p=0.0482$ for HFEBG, $p<0.01$ for GP, and $p=0.0142$ for random search). In particular, the comparison with random search directly addresses whether the LLM merely generates benchmarks at random: since random search reaches the theoretical minimum in only 3 of 10 trials whereas LLM-EBG reaches it in 8 of 10, the LLM-based operators are shown to exploit the search history conveyed through the in-context examples rather than acting as a random generator. The GP baseline performs worst of all, failing to reach the theoretical minimum in any trial. Here, it should be noted that, since the population size and the number of generations are matched to those of LLM-EBG, the number of benchmark evaluations ($N \times g_{max}$) is considerably smaller than the settings commonly adopted for GP. Under such a small evaluation budget, the GP baseline is unable to explore the search space sufficiently in the more challenging GA-preferred scenario, and its best expressions tend to remain structurally simple (e.g., constant-valued or single-term expressions). We therefore interpret this result as showing that, under an evaluation budget identical to that of LLM-EBG, the GP baseline is markedly less effective at constructing landscapes on which GA outperforms DE, rather than as evidence that GP is inherently incapable of generating such benchmarks.
 
In contrast, in the DE-preferred scenario, the GP and random search baselines both reach the theoretical minimum in all 10 trials, and no significant difference from LLM-EBG is observed ($p=0.647$ for both). This is because DE generally outperforms GA on unconstrained single-objective continuous problems, so landscapes favoring DE are easy to obtain and even a random generation of symbolic expressions attains the theoretical minimum. Consequently, the DE-preferred scenario is not discriminative among the symbolic-expression-based methods, and the advantage of LLM-EBG is instead demonstrated in the more challenging GA-preferred scenario. HFEBG is the only method that fails in the DE-preferred scenario ($p<0.01$), reflecting the limited diversity of its parameterized Gaussian representation, which makes it difficult to produce benchmarks that favor DE.
 
Beyond the overall performance comparison above, a further notable difference between the methods appears in whether the generated benchmarks satisfy the required problem dimensionality. In this study, the target is to generate five-dimensional benchmark problems; that is, a valid benchmark should depend on all five design variables. Importantly, neither LLM-EBG nor the GP baseline explicitly enforces the use of every variable: the LLM prompt only states that a five-dimensional problem should be generated, and the GP primitive set merely makes all five variables available. Under these conditions, LLM-EBG and the random search baseline, both of which rely on the LLM to generate expressions, produce benchmarks that incorporate all five design variables in every trial. In contrast, the best expressions generated by the GP baseline contain only $1.6$ variables on average in the GA-preferred scenario and $2.3$ in the DE-preferred scenario, and in three of the ten GA-preferred trials the GP baseline produced constant expressions that contain no design variables at all (e.g., \texttt{abs(sinh(2))}). Thus, the GP baseline frequently fails to generate benchmarks that meet the required five-dimensionality, effectively producing problems of lower dimensionality than intended. This contrast indicates that, without explicit instruction, the LLM interprets the requested dimensionality from the prompt and consistently generates problems that involve all design variables, which is a clear advantage of LLM-EBG over the GP baseline.
\color{black}

\subsection{Landscape Characteristics of Generated Benchmarks}\label{sec:analysis_character}
This section analyzes the characteristics of the generated benchmark problems using Exploratory Landscape Analysis (ELA)~\cite{Kerschke2019}. ELA is a methodology for estimating the geometrical properties of optimization landscapes, including convexity, objective-value distribution, level sets, curvature, and the fitness of linear and quadratic regression models. These features are computed from datasets sampled using Latin hypercube sampling (LHS). By employing ELA, we aim to quantitatively characterize the structural differences between the benchmarks generated in the GA-preferred and DE-preferred searches.

\begin{figure}[tb]
    \centering
    \includegraphics[width=\linewidth]{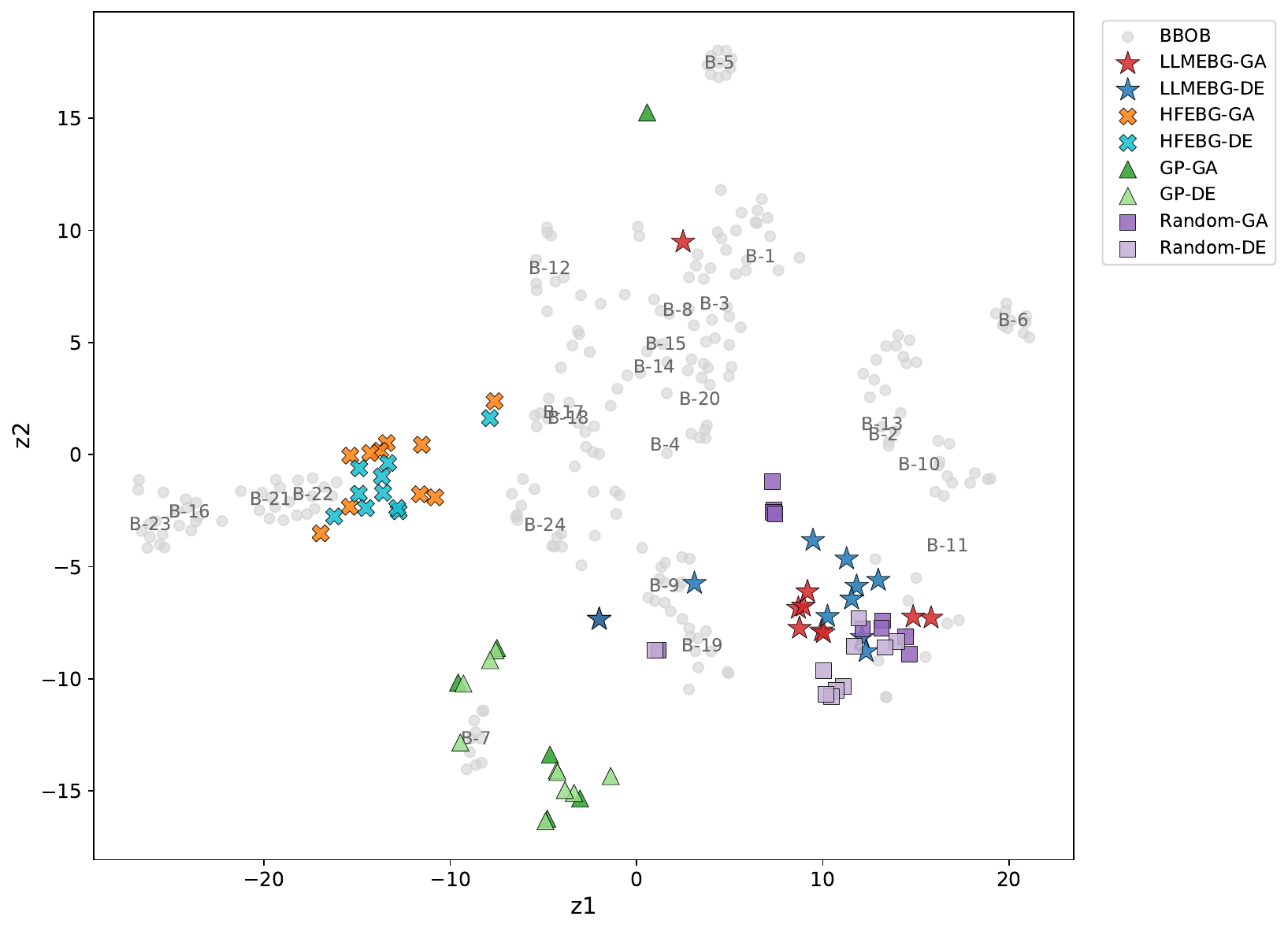}
    \caption{\modifiedsnd{Two-dimensional visualization of ELA features of benchmark problems generated by LLM-EBG and baseline methods with BBOB benchmark problems}}
    \label{fig:t-SNEofELA}
\end{figure}
Fig.~\ref{fig:t-SNEofELA} presents a two-dimensional embedding obtained by t-Distributed Stochastic Neighbor Embedding (t-SNE)~\cite{maaten2008visualizing} based on the ELA features extracted from both the generated benchmark problems and the BBOB benchmarks~\cite{Hansen2010} included in COCO. This visualization provides a qualitative overview of the relative positioning of the generated benchmarks within the ELA feature space. \modifiedsnd{Because the benchmarks generated by the GP baseline yield near-flat landscapes for which the curvature-based ELA features (i.e., the gradient- and Hessian-related features) are undefined, these features are excluded for all methods when computing the embedding, and the remaining ELA features are used.}

From Fig.~\ref{fig:t-SNEofELA}, we can find that many of the benchmarks \modified{generated by LLM-EBG} are embedded in regions close to BBOB10, BBOB11, and BBOB19. BBOB10 (Ellipsoidal Function) and BBOB11 (Discus Function) are predominantly quadratic functions characterized by strong variable scaling, where one or a few directions in the search space dominate the landscape geometry. In contrast, BBOB19 (Composite Griewank-Rosenbrock Function F8F2) exhibits pronounced multi-modality combined with narrow curved valleys, similar to those observed in the Rosenbrock function. The proximity of the generated benchmarks to these functions suggests that LLM-EBG can produce landscapes that combine anisotropy and nonlinearity, which are known to challenge optimization algorithms.

\modified{Compared with HFEBG, LLM-EBG and HFEBG are classified into different clusters, indicating that the benchmark problems generated by the two methods possess different structural characteristics. In particular, the benchmarks generated by HFEBG form a tight cluster in the t-SNE space near BBOB21 and BBOB22. BBOB21 (Gallagher's Gaussian 101-me Peaks Function) and BBOB22 (Gallagher's Gaussian 21-hi Peaks Function) consist of 101 and 21 optima with randomly chosen positions and heights, respectively. The proximity of HFEBG-generated benchmarks to these functions is consistent with the nature of HFEBG, which explicitly generates benchmark problems with $n_0 = 30$ local optima by searching within the parameter space of a fixed problem generator. This suggests that HFEBG tends to produce problems with structural characteristics similar to multimodal functions with a fixed number of local optima. In contrast, while many of the benchmarks generated by LLM-EBG form a cluster in a similar region of the t-SNE space, some trials produce problems with distinctly different landscape characteristics, spanning a broader area than those generated by HFEBG. This indicates that LLM-EBG is capable of generating a more diverse set of benchmark landscapes than HFEBG, owing to the representational flexibility afforded by LLMs.}

\modifiedsnd{The GP and random search baselines exhibit contrasting behaviors in the same feature space. The benchmarks generated by the random search baseline overlap with those of LLM-EBG, occupying a similar region of the embedding. This is expected because both methods generate benchmarks with the same LLM-based generator, so their raw structural distributions are similar; the difference between them lies not in the attainable landscape space but in how effectively that space is exploited under the guidance of the evaluation function, as reflected in the performance gap reported in Section~\ref{sec:llm-ebg-performance}. In contrast, the benchmarks generated by the GP baseline form a distinct cluster that is clearly separated from those of all other methods and is located closest to BBOB7. BBOB7 (Step-Ellipsoid Function) is characterized by many plateaus, over which the gradient is zero almost everywhere except in a small region near the global optimum. The proximity of the GP-generated benchmarks to BBOB7 is consistent with the tendency of the GP baseline, under the evaluation budget matched to LLM-EBG, to produce structurally simple, near-flat expressions (e.g., constant-valued or single-term expressions); indeed, the curvature-based ELA features of these benchmarks are undefined precisely because their landscapes are almost everywhere flat. Such plateau-dominated landscapes provide little structural information to distinguish the two algorithms, which is consistent with the failure of the GP baseline to generate GA-favoring benchmarks in the GA-preferred scenario under this small evaluation budget. We emphasize that this observation characterizes the behavior of the GP baseline under the same budget as LLM-EBG, and does not by itself imply an inherent limitation of GP.}

\begin{figure}[tb]
    \centering
    \includegraphics[width=\linewidth]{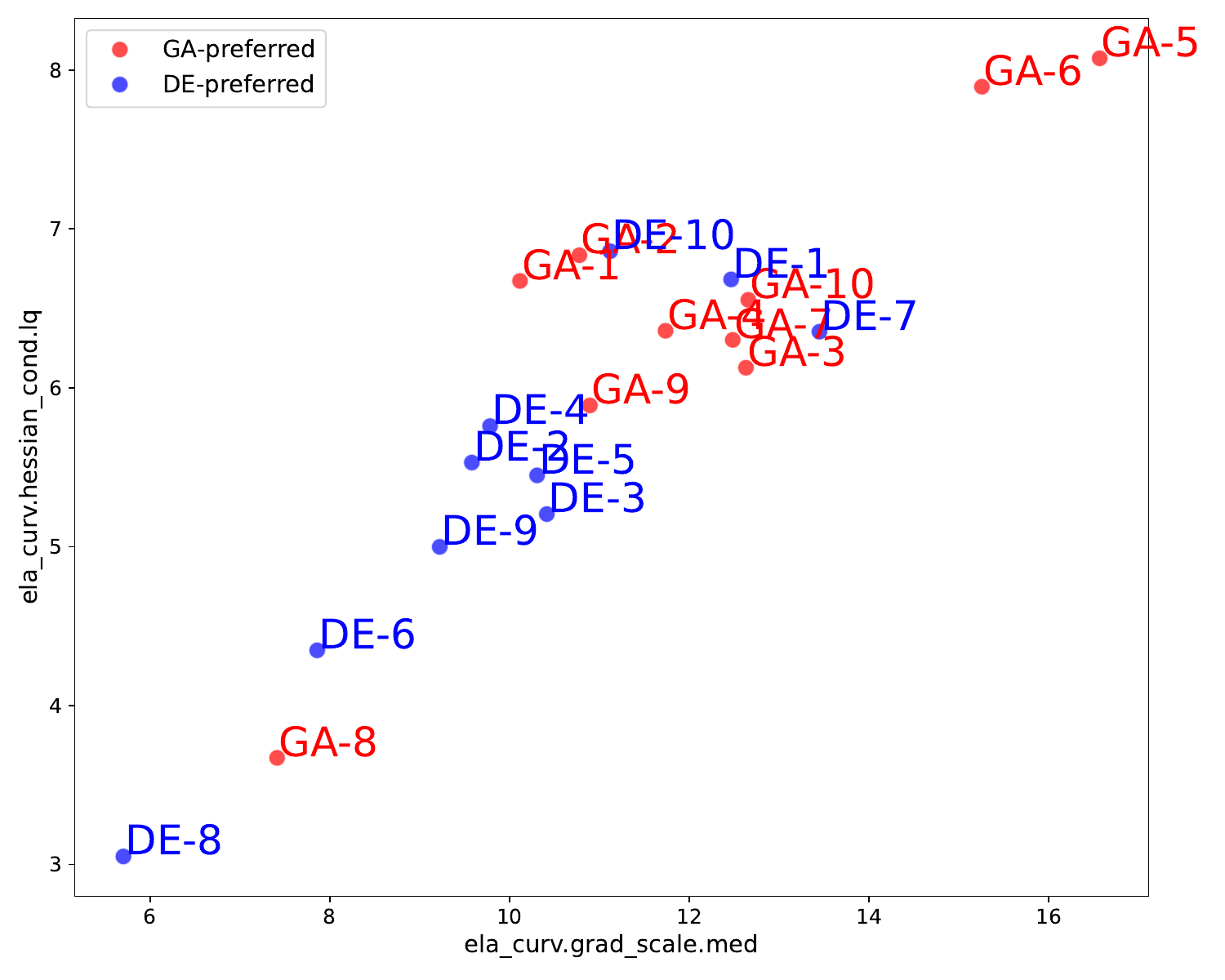}
    \caption{Distribution ELA features corresponding to the first and third largest loadings of the first principal component}
    \label{fig:elaPCA}
\end{figure}
Fig.~\ref{fig:elaPCA} shows the results of a principal component analysis (PCA) performed on the ELA features of the generated benchmark problems. The scatter plot visualizes the features associated with the largest and third-largest loadings of the first principal component (PC1). The third-largest loading is selected for the vertical axis because the second-largest loading corresponds to a feature belonging to the same category as the largest one. Specifically, the horizontal axis represents the \emph{median of the ratio of maximum to minimum absolute partial derivatives}, while the vertical axis represents the \emph{lower quartile of the condition number of the ratio of maximum to minimum Hessian matrix values}.

As shown in Fig.~\ref{fig:elaPCA}, benchmarks generated by the GA-preferred search tend to exhibit higher values for both the median partial-derivative ratios and the lower quartile of the Hessian condition numbers compared to those generated by the DE-preferred search. This indicates that the benchmarks favored by GA are more strongly influenced by variable scaling, corresponding to ill-conditioned landscapes with large differences in curvature across dimensions and locally elongated, elliptical level sets. The effectiveness of GA on these benchmarks can be explained by its ability to maintain population diversity and explore the search space robustly under strongly anisotropic conditions. 

In contrast, benchmarks generated by the DE-preferred search generally exhibit lower sensitivity to variable scaling and more moderate curvature disparities. Such landscapes provide clearer and more consistent searhch directions, allowing DE to exploit relative positional differences between individuals more effectively. This observation is consistent with the well-known tendency of DE to perform favorably on problems with smoother landscapes and less severe ill-conditioning.

\subsection{Representative Generated Benchmarks}
Fig.~\ref{fig:bestBenchmarks} presents two representative benchmark problems, GA-preferred-5 and DE-preferred-8, which exhibit markedly different characteristics in Fig.~\ref{fig:elaPCA}. Although this section shows representative results from the 10 trials, the full set of generated benchmark problems is provided in the supplementary material.

\begin{figure}[tb]
\begin{minipage}[t]{\linewidth}
\begin{tcolorbox}
\small
$f(x)=x_0^2 + \sin(x_1)x_2 + |x_3-x_4| + \sqrt{|x_0-x_1|} + \dfrac{x_2x_3}{1+x_4^2 + |\sin(x_0)\sinh(x_1)|} + \sinh(x_0)\cos^2(x_1) + \dfrac{|x_2-x_3|^2}{1+|x_4|} + \dfrac{x_0x_1x_2x_3x_4}{1+\sum_{i=0}^{4}|x_i|} + \sin(x_0)\sin(x_1)\sin(x_2)\sin(x_3)\sin(x_4) + \dfrac{|x_0-x_1|^2}{1+x_2^2} + \cos(x_0)\cos(x_1)\cos(x_2)\cos(x_3)\cos(x_4) + \dfrac{x_3x_4}{1+|x_0|+|x_1|+|x_2|}$
\end{tcolorbox}
\subcaption{GA-preferred-5}
\label{fig:ga-5-formula}
\end{minipage}
\begin{minipage}[t]{\linewidth}
\begin{tcolorbox}
\small
$f(x)=x_0^2 + |x_1x_2| + \sqrt{|x_3|} - \sin(x_4) + \sin(x_0x_1) + \cos(x_2x_3) + \dfrac{x_0}{1+x_4^2} + \sinh(x_1x_2x_3) + |x_0-x_1+x_2-x_3+x_4| + \sqrt{x_0^2+x_1^2+x_2^2+x_3^2+x_4^2} + x_1\sinh(x_0x_2) + \dfrac{|x_2-x_3|}{\sqrt{1+x_4^2}}
$
\end{tcolorbox}
\subcaption{DE-preferred-8}
\label{fig:de-8-formula}
\end{minipage}
\caption{Representative benchmarks generated in GA-preferred and DE-preferred searches}
\label{fig:bestBenchmarks}
\end{figure}

A comparison of the benchmarks generated in the GA-preferred and DE-preferred searches in Fig.~\ref{fig:bestBenchmarks} shows a clear structural contrast.
The GA-preferred search predominantly produces expressions involving trigonometric and power functions.
In contrast, the DE-preferred search tends to generate expressions characterized by more intricate interactions among variables.

In both scenarios, the generated benchmarks are explicitly constructed to avoid undefined numerical values, such as complex numbers or Not-a-Number (NaN) results. 
Specifically, radicands in square-root terms are enforced to be nonnegative through absolute-value or squared expressions. 
Similarly, denominators in fractional terms are expressed as sums of positive constants and absolute-value or squared terms, thereby preventing division by zero. 
\modified{These structural characteristics are not enforced by the system through explicit constraints. Rather, they are presumed to stem from the mathematical knowledge acquired by LLMs during pre-training. Specifically, mathematical expressions that appear in training corpora are typically written to avoid computationally infeasible regions such as division by zero and complex numbers, as such mathematically well-formed expressions are prevalent in the training data. As a result, LLMs naturally tend to generate expressions with these properties without being explicitly instructed to do so. This suggests that by leveraging LLMs, it is possible to generate problems that do not include computationally infeasible regions without explicitly defining such constraints in advance.}

\begin{figure}[tb]
    \centering
     \begin{minipage}[t]{0.48\columnwidth}
    \includegraphics[width=0.9\linewidth]{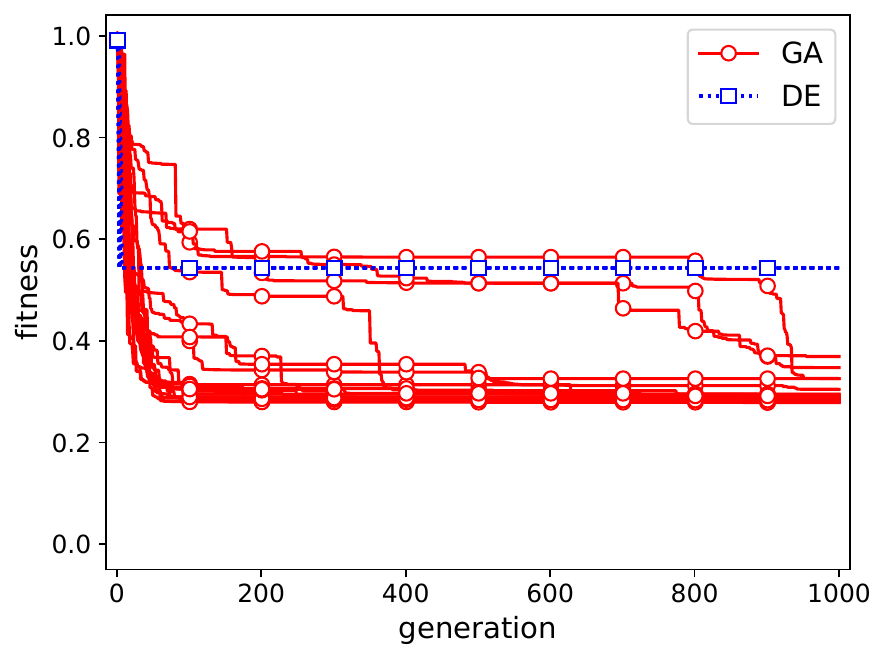}
     \subcaption{GA-preferred-5}
\label{fig:EAresult_GApreferred}
     \end{minipage}
    \hfill
    \begin{minipage}[t]{0.48\columnwidth}
    \includegraphics[width=0.9\linewidth]{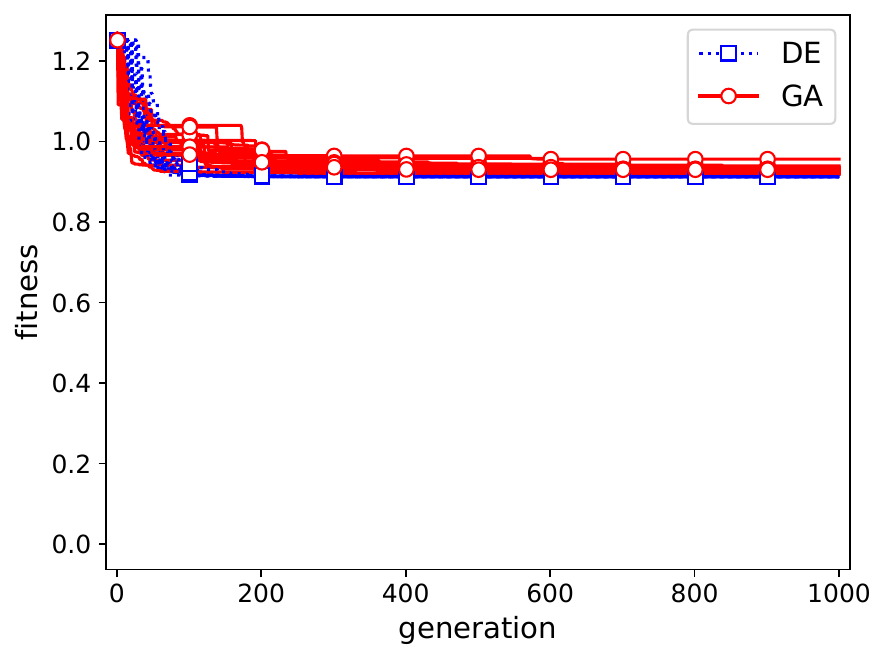}
    \subcaption{DE-preferred-8}
\label{fig:EAresult_DEpreferred}
    \end{minipage}
    \caption{Search trajectories of GA and DE on representative generated benchmarks}
    \label{fig:EAresult}
\end{figure}
Fig.~\ref{fig:EAresult} illustrates the search processes of GA and DE on the representative benchmarks shown in Fig.~\ref{fig:bestBenchmarks}. In Fig.~\ref{fig:EAresult}, the horizontal axis represents the number of generations, and the vertical axis denotes the objective function value. The red solid and blue dotted curves correspond to the search trajectories of GA and DE, respectively.

As shown in Fig.~\ref{fig:EAresult_GApreferred}, for the benchmark generated in the GA-preferred search, DE tends to converge to specific local optima in early generations, whereas GA consistently avoids premature convergence and identifies superior solutions across all trials. Conversely, for the benchmark generated in the DE-preferred search (Fig.~\ref{fig:EAresult_DEpreferred}), DE converges rapidly toward near-optimal solutions, while GA exhibits convergence to multiple local optima.

A comparison between the two algorithms indicates that GA exhibits greater variability in the best objective values across trials but can escape local optima, whereas DE exhibits a strong tendency toward early convergence and is comparatively more susceptible to stagnation at local optima.

These observations indicate that LLM-EBG can generate benchmark problems that induce substantial performance disparities between GA and DE, and that the structural characteristics of the generated benchmarks reflect the distinct convergence behaviors of the two algorithms.

To further clarify how individual design variables influence the generated landscapes, a Sobol' global sensitivity analysis~\cite{Sobol1993} was conducted. 
\modified{Sobol' global sensitivity analysis is a variance-based method that decomposes the output variance of a model into contributions attributable to each input variable and their interactions. Specifically, it computes two types of sensitivity indices: the first-order sensitivity index ($S_1$), which measures the direct contribution of each variable to the output variance, and the total-order sensitivity index ($S_T$), which accounts for the variable's direct effect as well as all higher-order interactions with other variables. A higher sensitivity index indicates a greater influence of the corresponding variable on the objective function value of the benchmark problem.}

\begin{figure}[tb]
\centering
\begin{minipage}[t]{0.48\linewidth}
\centering
\includegraphics[width=\linewidth]{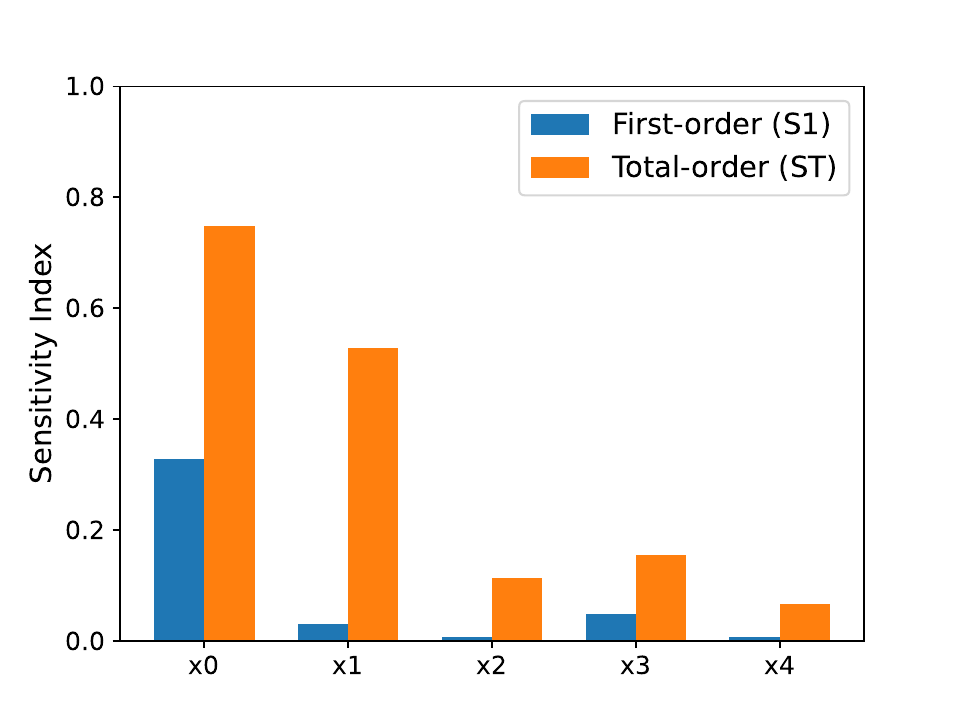}
\subcaption{GA-preferred-5}
\end{minipage}
\hfill
\begin{minipage}[t]{0.48\linewidth}
\centering
\includegraphics[width=\linewidth]{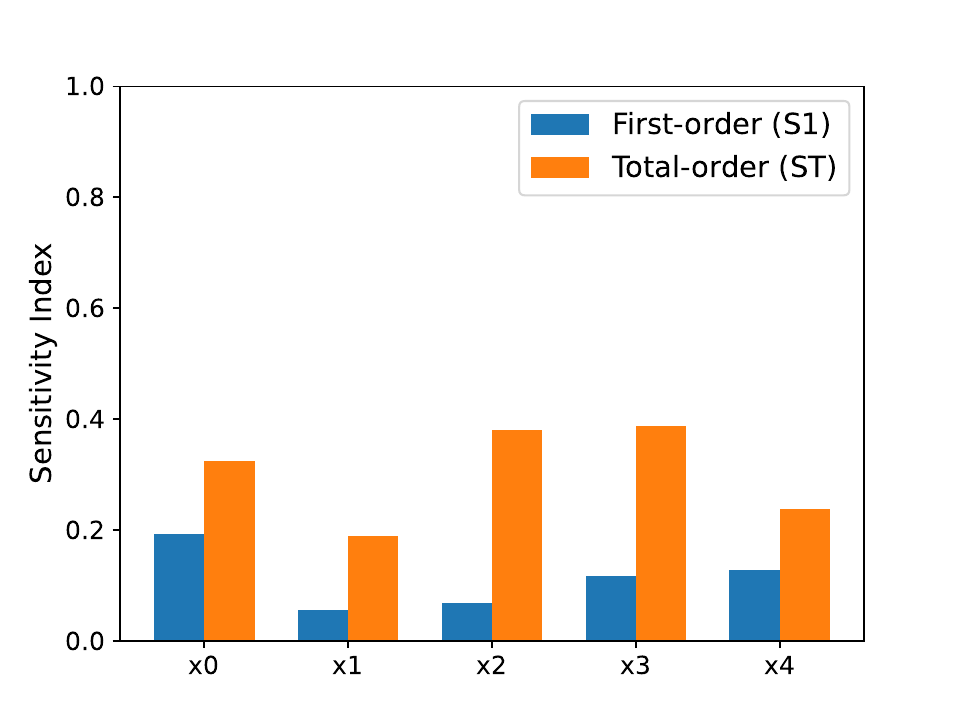}
\subcaption{DE-preferred-8}
\end{minipage}
\caption{Sobol’ sensitivity analysis for representative generated benchmarks}
\label{fig:sobol}
\end{figure}
Fig.~\ref{fig:sobol} presents the results of the Sobol' analysis for the representative benchmarks GA-preferred-5 and DE-preferred-8. The horizontal axis shows the five design variables, and the vertical axis shows the sensitivity indices, with the blue and orange bars representing the first-order and total-order sensitivity indices, respectively.

As shown in Fig.~\ref{fig:sobol}, for the benchmark generated in the GA-preferred search, the first-order sensitivity index of $x_0$ is larger than that of the remaining variables, indicating that $x_0$ has a dominant influence on the objective function. Moreover, the total-order sensitivity indices for $x_0$ and $x_1$ exceed those for $x_2$ to $x_4$, suggesting substantial interaction effects with other variables.

In contrast, the benchmark generated in the DE-preferred search exhibits a more uniform distribution of variable influence. Specifically, while the total variance of the objective function is $1.56$ for GA-preferred-5 and $2.27$ for DE-preferred-8, the variance contributions of $x_0$ are $0.510$ and $0.435$, respectively, and those of $x_1$ are $0.0479$ and $0.127$, respectively. These results are consistent with the landscape characteristics observed in Fig.~\ref{fig:elaPCA}.

Overall, these results indicate that LLM-EBG effectively explored benchmark landscapes with diverse variable scales, thereby facilitating the generation of benchmark problems that induce substantial performance disparities between GA and DE.

\section{Analysis of Evolutionary Operator Behavior in LLM-EBG}\label{sec:analysis_behavior}
For each scenario, we analyze the execution behavior of evolutionary operators during the optimization process.

\subsection{LLM-based Operator Statistics}
Table~\ref{tab:allFamilyree} summarizes statistics characterizing the amount and composition of LLM-based genetic operations required to reach the optimal benchmark instances. Specifically, we report the average number of ancestral individuals along the lineage leading to the optimal benchmark, the total number of genetic operations executed by the LLM along this lineage, and the ratio of crossovers to mutations. Cases in which the LLM produced offspring identical to their parents are excluded from the operator count.
\begin{table}[tb]
    \centering
    \caption{\modified{Statistics of LLM-based genetic operations for trials that reached the ideal fitness}}
    \label{tab:allFamilyree}
    \begin{tabular}{lccc}
    \toprule
    &GA-preferred&DE-preferred &$p$-value \\
    \midrule
    Average number of individuals&8.88&8.77&0.735\\
    Average number of genetic operations&6.75&6.78&0.734\\
    Ratio of crossover to mutation&51.9\%&45.0\%&0.329\\
    \bottomrule
    \end{tabular}
\end{table}


\modified{As shown in Table~\ref{tab:allFamilyree}, both scenarios exhibit broadly similar numbers of ancestral individuals and genetic operations to reach the optimal benchmark instances, despite the DE-preferred search reaching the theoretical minimum in fewer generations on average than the GA-preferred search, as reported in Section~\ref{sec:llm-ebg-performance}. However, Wilcoxon rank-sum tests confirm no statistically significant differences between the two scenarios in terms of the number of ancestral individuals ($p = 0.735$) and the number of genetic operations ($p = 0.734$). This suggests that the depth of structural transformations applied by the LLM tends to be comparable across the two scenarios. In addition, there is a possible trend that effective improvements to benchmark structures occur somewhat more readily when evolving DE-favoring landscapes.}

\modified{The DE-preferred search exhibits a crossover-to-mutation ratio that is slightly lower than that of the GA-preferred search and somewhat lower than the nominal crossover probability of 50\%, though a Wilcoxon rank-sum test confirms no statistically significant difference in the crossover-to-mutation ratio between the two scenarios ($p = 0.329$). This suggests that mutation operations tend to play a relatively more prominent role in refining benchmark structures in the DE-preferred search. In contrast, in the GA-preferred search, crossover and mutation appear to be utilized at broadly similar frequencies, suggesting that both recombination and local structural modification contribute comparably to the construction of GA-favoring benchmarks.}


\subsection{Analysis of LLM-based Genetic Operator Patterns}\label{sec:operatorPattern}

An analysis of the generation process reveals that LLM-EBG produces offspring by repeatedly applying four types of modifications to parent benchmark expressions. 
The observed mutation behaviors are classified into the following four types:
\begin{description}
\item[\textbf{Type I: Constant and index modification}:]
Modification of numerical constants and variable indices in the parent expression.
E.g.,
\begin{align*}
&\;\textbf{Parent:}\; f(x)=x_0^2+x_2^2+x_4\sin(x_3)\\
\rightarrow&\;\textbf{Offspring:}\; f(x)=x_0^2+x_2^3+x_2\sin(x_1)
\end{align*}

\item[\textbf{Type II: Functional substitution}:]
\modified{Substitution of functional terms in the parent expression.}
E.g.,
\begin{align*}
&\;\textbf{Parent:}\; f(x)=x_0x_1+\sin(x)\\
\rightarrow&\;\textbf{Offspring:}\; f(x)=x_0\cos(x_1)+\sinh(x)
\end{align*}

\item[\textbf{Type III: Variable deletion}:]
Deletion of variables or subexpressions in the parent expression.
E.g.,
\begin{align*}
&\;\textbf{Parent:}\; f(x)=\sqrt{x_0^2+x_1^2+x_3^2+x_4^2}\\
\rightarrow&\;\textbf{Offspring:}\; f(x)=\sqrt{x_0^2+x_1^2}
\end{align*}

\item[\textbf{Type IV: Novel term generation}:]
Generation of novel symbolic terms not present in the parent expression.
E.g.,
\begin{align*}
&\;\textbf{Parent:}\; f(x)=x_0+\sin(x_1)^2\\
\rightarrow&\;\textbf{Offspring:}\; f(x)=x_0+\sin(x_1)^2+x_0\sinh(x_1)
\end{align*}

\end{description}

Our analysis shows that offspring expressions tend to preserve the leading part of the parent expressions and apply additions, substitutions, or deletions primarily to the latter part of the formula. This tendency is particularly pronounced during crossover operations. It can be attributed to the structural similarity between the leading parts of the parent expressions, which induces selection pressure on the LLM to inherit core problem characteristics while modifying peripheral components.

Furthermore, the Type I and Type IV modifications were observed most frequently. \modified{Type I modifications may reflect the LLM's tendency to prefer conservative, local changes, while Type IV modifications may reflect its tendency to expand expressions. This behavioral bias restricts the diversity in the search for high-quality benchmark problems, and it indicates the need for improvement. Nevertheless, it is worth noting that LLM-based modifications differ fundamentally from random mutations in GP or standard EAs: unlike random mutations, which apply changes without considering the global structure of the expression, LLMs generate modifications while taking into account the overall mathematical context. This suggests that LLM-based operators can produce mathematically meaningful modifications more readily than purely random approaches.}


\subsection{Visualization and Structural Analysis of Evolutionary Lineages in LLM-EBG}

\begin{figure}[tb]
    \centering
    \begin{minipage}[t]{0.48\linewidth}
    \includegraphics[width=\linewidth]{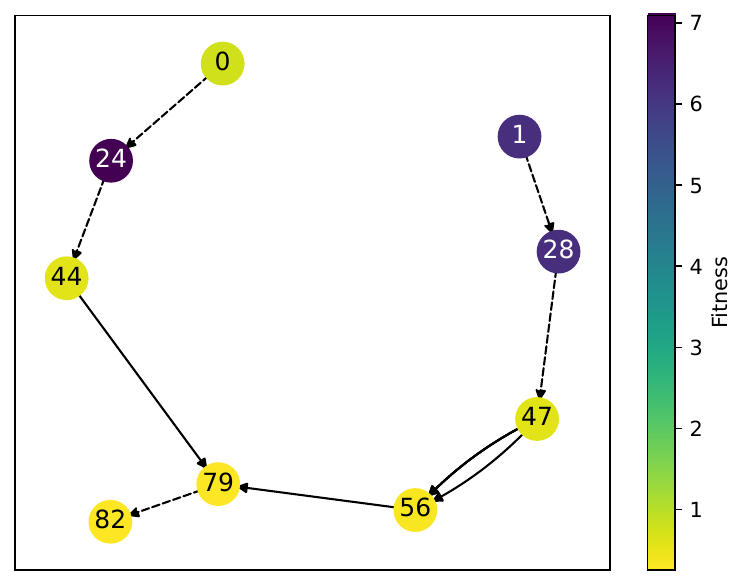} \subcaption{GA-preferred} \label{fig:networkBest_ga}
    \end{minipage}
    \hfill
    \begin{minipage}[t]{0.48\linewidth}
    \includegraphics[width=\linewidth]{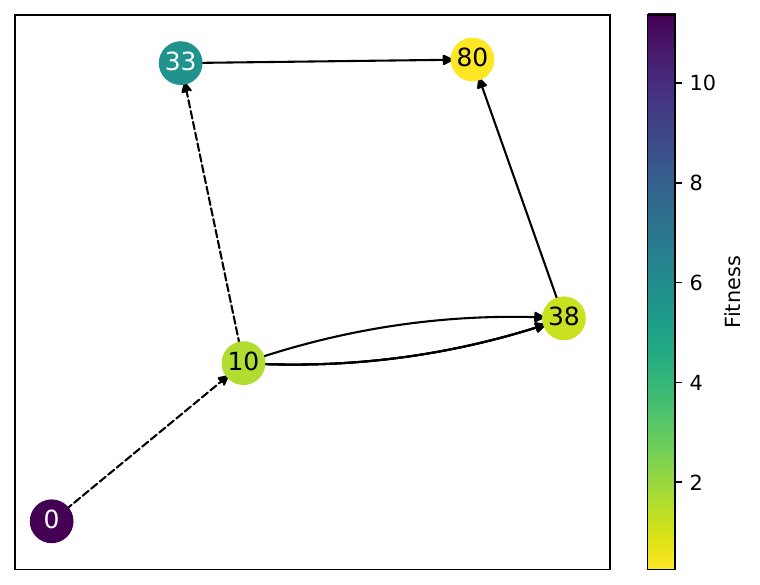} \subcaption{DE-preferred} \label{fig:networkBest_de}
    \end{minipage}
    \caption{Evolutionary lineages of benchmark generation in LLM-EBG. \modified{The spatial layout is determined by computing Levenshtein distances between benchmark expressions and embedding them into a two-dimensional space using multidimensional scaling (MDS).}}
    \label{fig:networkBest}
\end{figure}
Fig.~\ref{fig:networkBest} visualizes the evolutionary lineages that led to the generation of the optimal benchmark instances. The figure depicts the result of the trial whose number of generations is closest to the median across all runs. In these graphs, each node represents a generated benchmark problem, where the node index corresponds to its generation order and the node color indicates its fitness value. Directed edges denote parent–offspring relationships formed by LLM-based genetic operations; an edge from node $A$ to node $B$ indicates that $B$ was generated by applying a genetic operator to $A$. Solid edges represent crossover operations, whereas dotted edges represent mutation operations. The spatial layout is determined by computing Levenshtein distances between benchmark expressions and embedding them into a two-dimensional space using multi-dimensional scaling (MDS)~\cite{MDS}.

\modified{As shown in Fig.~\ref{fig:networkBest_ga}, in this representative trial, the GA-preferred search exhibits multiple evolutionary branches in the early and middle stages, where mutation operations drive structural exploration. In the later stages, an individual generated by crossing over two individuals from distinct lineages (nodes 44 and 56 producing node 79) contributes to reaching the optimal benchmark, suggesting that combining structural components from different lineages can be effective in the final stages of the search. It is worth noting, however, that some crossover operations involve the same parent individual (e.g., node 47), which functionally resembles mutation rather than recombination.}

\modified{In contrast, Fig.~\ref{fig:networkBest_de} shows that in this representative trial, the DE-preferred search exhibits a more linear evolutionary trajectory. Although two crossover operations are observed, one of them involves the same parent individual (node 10), which effectively functions as a mutation-like operation. This suggests that the DE-preferred search in this trial proceeds mainly through localized modifications within a single lineage, resembling a mutation-driven local search process.}

\modified{Across both scenarios, these observations suggest that the distinction between crossover and mutation in LLM-EBG may be more nuanced than in conventional evolutionary algorithms, as some crossover operations effectively function as mutation when the same parent is selected. This highlights a unique characteristic of LLM-based genetic operators that warrants further investigation.}
To further enrich structural diversity, providing the LLM with more explicit mutation patterns and transformation templates could be beneficial. Moreover, the GA-preferred search leverages information exchange across the population via crossover, whereas the DE-preferred search relies on mutation-centered, lineage-based exploration, highlighting two distinct evolutionary dynamics within the proposed framework.

\modified{
\section{Limitations of the Proposed Method}\label{sec:limitation}
This section discusses the limitations of LLM-EBG. By leveraging LLMs, it is possible to generate desired benchmark problems without requiring domain-specific knowledge, and the experimental results demonstrate that LLM-EBG can generate benchmarks that enable a comparative evaluation of GA and DE. On the other hand, LLMs require substantial computational time to generate equation instances, which leads to an increase in the overall execution time of the proposed method, as discussed in Section~\ref{sec:params}. This computational cost also limits the scalability of LLM-EBG to higher-dimensional problems: increasing the dimensionality of the design space either leads to an increase in the input prompt size or results in overly smoothed benchmark problems. Compared to GP-based approaches, which can generate benchmark expressions at a much lower computational cost, LLM-EBG requires significantly more time per generation due to the overhead of LLM queries. Therefore, it is necessary to adopt more efficient approaches for generating candidate problems, such as utilizing smaller-scale models and reducing prompt size. Furthermore, extending LLM-EBG to generate benchmarks represented as program code, rather than mathematical expressions, could potentially address the scalability issue, as program code can naturally handle arbitrary input dimensions through constructs such as loops and array operations, without requiring explicit enumeration of variables.
}

\section{Conclusion and Future Work}\label{sec:conclusion}
This study proposes LLM-EBG, an evolutionary framework for automatically generating benchmark problems using LLMs. As a case study, we focused on generating unconstrained single-objective continuous minimization problems and investigated whether LLM-EBG can generate benchmark instances that induce significant performance differences between two representative EAs, GA and DE. Specifically, we considered two scenarios: GA-preferred and DE-preferred searches.

Experimental results demonstrated that LLM-EBG successfully generated benchmark problems in which GA outperformed DE in 8 of 10 trials under the GA-preferred scenario, and in which DE outperformed GA in 9 of 10 trials under the DE-preferred scenario. These results indicate that the proposed framework can effectively steer the benchmark generation process toward landscapes that favor specific algorithmic behaviors.

Landscape analysis using ELA revealed that benchmarks generated in the GA-preferred search exhibit significantly higher sensitivity to variable scaling than those generated in the DE-preferred search. This finding suggests that LLM-EBG can automatically identify and amplify key geometric properties, such as anisotropy and ill-conditioning, that are known to differentiate the search behaviors of EAs.

Furthermore, analysis of the LLM-based evolutionary operator showed that DE-favoring benchmarks are generated more readily through mutation-dominated exploration, whereas GA-favoring benchmarks tend to emerge from a search process that actively leverages information exchange among individuals via crossover. These observations indicate that LLM-EBG not only generates benchmarks with desired performance characteristics but also exhibits algorithm-specific evolutionary dynamics.

Future work will extend the proposed framework to a broader range of optimization algorithms beyond GA and DE, and to more general problem classes, including constrained and multi-objective optimization. In addition, improving the diversity of the initial benchmark population remains an important challenge for enhancing the robustness and exploratory capability of LLM-EBG. \modified{Furthermore, while this study focuses on generating benchmarks represented as mathematical expressions that emphasize performance disparities between two algorithms, extending LLM-EBG to generate diverse benchmark suites---whether expressed as program code, independent of specific algorithm performance, or reflecting the complexity of real-world problems---represents an important direction for future research. It is also necessary to investigate the influence of input prompts on LLMs, specifically whether information on desirable problems or their evaluation values should be included.} Addressing these issues will further strengthen the potential of LLM-EBG as a general-purpose framework for automatic benchmark generation.

\section*{Statements and Declarations}

\paragraph{Funding}
This work was supported by the Joint Usage/Research Center for Interdisciplinary Large-scale Information Infrastructures (JHPCN) in Japan (Project ID: jh250063) and JSPS Grant-in-Aid for Scientific Research (B) (Grant Number: 24K03011).

\paragraph{Competing Interests}
The authors have no competing interests to declare that are relevant to the content of this article.

\paragraph{Availability of data and materials}
The datasets generated during and/or analyzed during the current study are available from the corresponding author upon reasonable request.

\paragraph{Code availability}
The source code used in this study is available from the corresponding author upon reasonable request.

\paragraph{Authors’ contributions}
\textbf{Yuhiro Ono: } Writing -- original draft, Conceptualization, Investigation, Methodology, Formal analysis, Software, Visualization. 
\textbf{Tomohiro Harada: } Writing -- review \& editing, Conceptualization, Methodology, Validation, Supervision, Funding Acquisition. 
\textbf{Yukiya Miura} Writing -- review \& editing, Supervision, Resource.

\paragraph{Declaration of Generative AI and AI-assisted technologies in the writing process}
During the preparation of this work, the authors used Word Processing tools, Spreadsheets, and AI assistants in order to improve the readability of the text and figures. After that, the authors reviewed and edited the content as needed and then took full responsibility for the published article.

\paragraph{Ethics approval}
Not applicable.

\paragraph{Consent to participate}
Not applicable.

\paragraph{Consent for publication}
Not applicable.

\begin{appendices}
\color{\modcolorsnd}
\section{Detailed Benchmark Generation Results for the Baseline Methods}\label{sec:append}
This appendix reports the results of the three baseline methods. Table~\ref{tb:resultHFEBG} summarizes the benchmarks generated by HFEBG, following the same structure as Table~\ref{tb:resAllSearch} except for the fitness values, which are defined as described in Section~\ref{sec:result}. Tables~\ref{tb:resultGP} and~\ref{tb:resultRandom} summarize the results of the GP and random search baselines, respectively, following the same structure as Table~\ref{tb:resAllSearch}. The fitness values of the GP and random search baselines are computed by the same evaluation function as LLM-EBG. 
 
\begin{table}[tb]
    \centering
    \caption{Best benchmarks obtained by HFEBG}
    \label{tb:resultHFEBG}
        \begin{tabular}{ccccc}
        \toprule
        Scenario&Trial&$fitness$&Best GA value&Best DE value\\
        \midrule
        \multirow{10}{*}{GA-preferred}&1&$0.261$&$-0.990$&$-0.751$\\
        &2&$0.280$&$-0.968$&$-0.968$\\
        &3&$0.720$&$-1.001$&$-1.001$\\
        &4&$\mathbf{0.256}$&$-0.862$&$-0.809$\\
        &5&$\mathbf{0.256}$&$-0.967$&$-0.948$\\
        &6&$0.285$&$-0.965$&$-0.899$\\
        &7&$\mathbf{0.256}$&$-1.032$&$-0.704$\\
        &8&$\mathbf{0.256}$&$-1.039$&$-0.961$\\
        &9&$0.268$&$-1.003$&$-0.796$\\
        &10&$0.695$&$-0.966$&$-0.966$\\
        \midrule
        \multirow{10}{*}{DE-preferred}&1&$0.280$&$-0.895$&$-0.932$\\
        &2&$0.350$&$-1.067$&$-1.068$\\
        &3&$0.548$&$-1.005$&$-1.008$\\
        &4&$0.590$&$-0.988$&$-1.015$\\
        &5&$0.410$&$-0.881$&$-0.923$\\
        &6&$0.351$&$-1.054$&$-1.083$\\
        &7&$0.300$&$-0.981$&$-0.852$\\
        &8&$0.468$&$-1.038$&$-1.047$\\
        &9&$0.354$&$-1.044$&$-1.044$\\
        &10&$0.437$&$-0.989$&$-1.004$\\
        \bottomrule
        \end{tabular}
\end{table}
 
\begin{table}[tb]
    \centering
    \caption{Best benchmarks obtained by the GP baseline}
    \label{tb:resultGP}
        \begin{tabular}{ccccc}
        \toprule
        Scenario&Trial&$fitness$&Best GA value&Best DE value\\
        \midrule
        \multirow{10}{*}{GA-preferred}&1&$0.500$&$1$&$1$\\
        &2&$0.500$&$3.30\times10^{41}$&$3.30\times10^{41}$\\
        &3&$0.402$&$0.00267$&$1.66\times10^{-37}$\\
        &4&$0.500$&$0.00$&$0.00$\\
        &5&$0.330$&$2.06\times10^{-5}$&$0.174$\\
        &6&$0.280$&$0.824$&$0.824$\\
        &7&$0.500$&$0.810$&$0.810$\\
        &8&$0.500$&$3.63$&$3.63$\\
        &9&$0.488$&$0.00$&$0.00$\\
        &10&$0.459$&$9.97\times10^{-8}$&$4.44\times10^{-195}$\\
        \midrule
        \multirow{10}{*}{DE-preferred}&1&$\mathbf{0.256}$&$0.00101$&$1.58\times10^{-81}$\\
        &2&$\mathbf{0.256}$&$0.0288$&$4.37\times10^{-41}$\\
        &3&$\mathbf{0.256}$&$4.99\times10^{-4}$&$6.14\times10^{-83}$\\
        &4&$\mathbf{0.256}$&$0.230$&$0.230$\\
        &5&$\mathbf{0.256}$&$3.58\times10^{-14}$&$1.19\times10^{-214}$\\
        &6&$\mathbf{0.256}$&$3.638$&$3.637$\\
        &7&$\mathbf{0.256}$&$1.65\times10^{-12}$&$1.38\times10^{-213}$\\
        &8&$\mathbf{0.256}$&$3.639$&$3.637$\\
        &9&$\mathbf{0.256}$&$1.10\times10^{-10}$&$5.55\times10^{-185}$\\
        &10&$\mathbf{0.256}$&$5.15\times10^{-6}$&$1.66\times10^{-67}$\\
        \bottomrule
        \end{tabular}
\end{table}
 
\begin{table}[tb]
    \centering
    \caption{Best benchmarks obtained by the random search baseline}
    \label{tb:resultRandom}
        \begin{tabular}{ccccc}
        \toprule
        Scenario&Trial&$fitness$&Best GA value&Best DE value\\
        \midrule
        \multirow{10}{*}{GA-preferred}&1&$\mathbf{0.256}$&$0.00135$&$0.239$\\
        &2&$0.646$&$0.296$&$0.296$\\
        &3&$\mathbf{0.256}$&$1.30\times10^{-4}$&$0.115$\\
        &4&$0.329$&$0.0201$&$0.0201$\\
        &5&$0.306$&$0.0202$&$0.0201$\\
        &6&$0.744$&$1$&$1$\\
        &7&$0.276$&$0.0202$&$0.0279$\\
        &8&$\mathbf{0.256}$&$0.557$&$0.592$\\
        &9&$0.261$&$1.78\times10^{-4}$&$0.239$\\
        &10&$0.280$&$0.0202$&$0.0201$\\
        \midrule
        \multirow{10}{*}{DE-preferred}&1&$\mathbf{0.256}$&$9.44\times10^{-8}$&$1.09\times10^{-115}$\\
        &2&$\mathbf{0.256}$&$1$&$1$\\
        &3&$\mathbf{0.256}$&$0.3228$&$0.3227$\\
        &4&$\mathbf{0.256}$&$1.98\times10^{-4}$&$5.34\times10^{-54}$\\
        &5&$\mathbf{0.256}$&$0.00121$&$7.99\times10^{-11}$\\
        &6&$\mathbf{0.256}$&$1$&$1$\\
        &7&$\mathbf{0.256}$&$1$&$1$\\
        &8&$\mathbf{0.256}$&$4.26\times10^{-4}$&$2.41\times10^{-32}$\\
        &9&$\mathbf{0.256}$&$1.39\times10^{-8}$&$4.54\times10^{-59}$\\
        &10&$\mathbf{0.256}$&$0.3228$&$0.3227$\\
        \bottomrule
        \end{tabular}
\end{table}
\color{black}
\end{appendices}
\bibliography{reference}

\end{document}